\documentclass{article}

\usepackage[numbers,sort,compress]{natbib} 
\usepackage{subcaption}
\usepackage{amsmath}
\usepackage{pifont} 
\usepackage{xcolor}
\newcommand{\cmark}{\textcolor{green!60!black}{\ding{51}}}  
\newcommand{\xmark}{\textcolor{red}{\ding{55}}}             
\usepackage{graphicx}
\usepackage{subcaption}
\usepackage{float} 
\usepackage{caption}
\usepackage{adjustbox}

\usepackage[preprint]{neurips_2025}
\newcommand{\horizontal}{\ensuremath{\mathtt{horizontal\:position\,}}}

\newcommand{\vertical}{\ensuremath{\mathtt{vertical\:position\,}}}
\newcommand{\depth}{\ensuremath{\mathtt{depth\,}}}
\newcommand{\proximity}{\ensuremath{\mathtt{proximity\,}}}
\newcommand{\order}{\ensuremath{\mathtt{order\,}}}

\usepackage[utf8]{inputenc} 
\usepackage[T1]{fontenc}    
\usepackage{hyperref}       
\usepackage{url}            
\usepackage{booktabs}       
\usepackage{amsfonts}       
\usepackage{nicefrac}       
\usepackage{microtype}      
\usepackage{xcolor}         

\usepackage{enumitem }

\usepackage{soul}
\usepackage{xcolor}

\newcommand{\hlgreen}[1]{{\sethlcolor{green!30}\hl{#1}}}
\newcommand{\hlorange}[1]{{\sethlcolor{orange!30}\hl{#1}}}

\title{Understanding Space Is Rocket Science - Only Top Reasoning Models Can Solve Spatial Understanding Tasks}

\author{%
  Nils~Hoehing \\
  School of Computer Science\\
  University College Dublin\\
  Dublin 4, Ireland \\
  \texttt{nils.hohing@ucdconnect.ie} \\
  \And
  Mayug Maniparambil \\
  School of Computing \\
  Dublin City University  \\
  Dublin 9, Ireland  \\
  \texttt{mayugmaniparambil@gmail.com} \\
  \And
  Ellen Rushe \\
  School of Computing \\
  Dublin City University\\
  Dublin 9, Ireland \\
  \texttt{ellen.rushe@dcu.ie} \\
  \And
  Noel E. O'Connor \\
  School of Electronic Engineering \\
  Dublin City University\\
  Dublin 9, Ireland \\
  \texttt{noel.oconnor@dcu.ie} \\
  \And
  Anthony Ventresque \\
  School of Computer Science and Statistics \\
  Trinity College Dublin \\
  Dublin 2, Ireland \\
  \texttt{anthony.ventresque@tcd.ie} \\
}

\begin{document}

\maketitle

\begin{abstract}
    We propose RocketScience, an open-source contrastive VLM benchmark that tests for spatial relation understanding. It is comprised of entirely new real-world image-text pairs covering mostly relative spatial understanding and the order of objects. The benchmark is designed 
    to be very easy for humans and hard for the current generation of VLMs, and this is empirically verified. Our results show a striking lack of spatial relation understanding in open source and frontier commercial VLMs and a surprisingly high performance of reasoning models. Additionally, we perform a disentanglement analysis to separate the contributions of object localization and spatial reasoning in chain-of-thought-based models and find that the performance on the benchmark is bottlenecked by spatial reasoning and not object localization capabilities. 
    We release the dataset with a CC-BY-4.0 license and make the evaluation code available at: \href{https://github.com/nilshoehing/rocketscience}{https://github.com/nilshoehing/rocketscience}.

\end{abstract}
\section{Introduction}
Language models with vision capabilities have enabled powerful applications \cite{vlmsurvey}, from turning sketches into website prototypes to recommending recipes based on a photo of fridge contents. At the same time, these models still struggle with fundamental tasks that are often trivial to humans, particularly those that centre around understanding spatial relationships between objects in an image.

\begin{figure}[t]
\includegraphics[width=\textwidth]{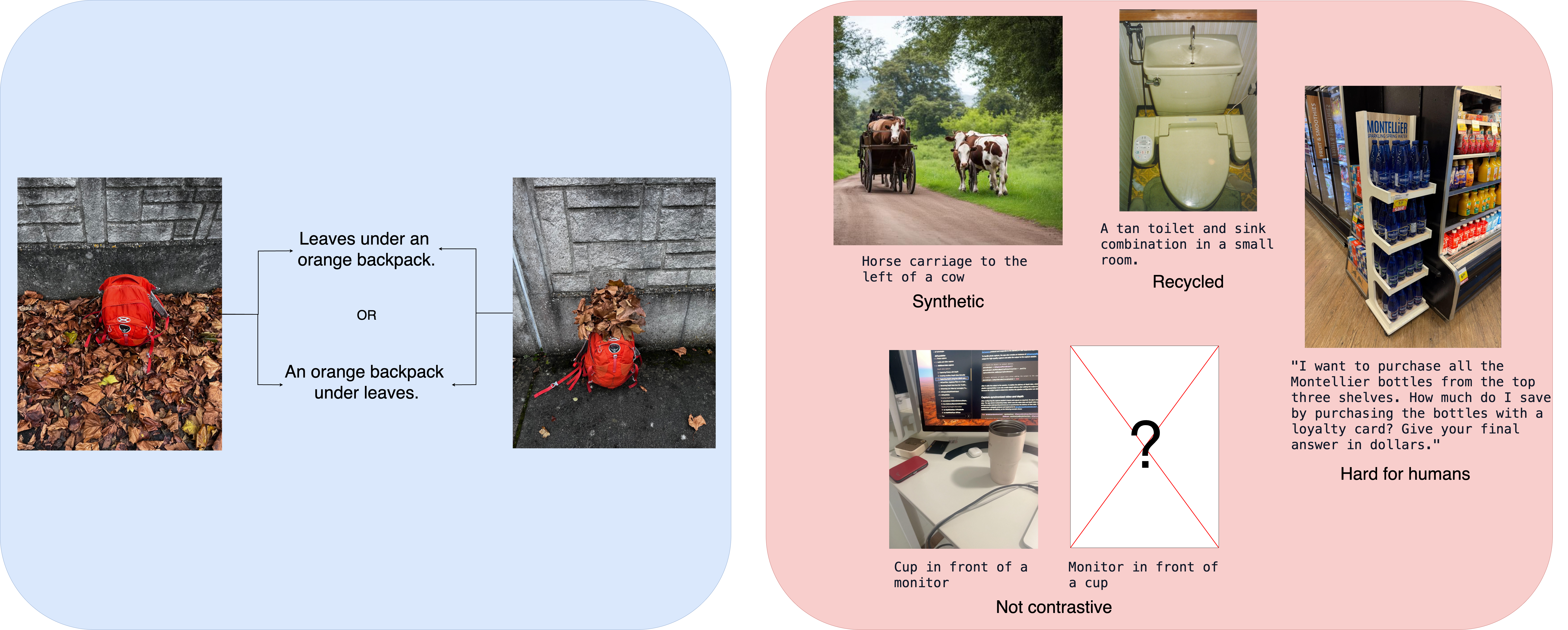}
\caption{Left: A contrastive pair from our RocketScience benchmark, showing two images and captions differing only in object positions. Right: Examples of problematic data in other benchmarks, such as reused or synthetic data, from \cite{vismin} (cc-by-sa-4.0), \cite{sc++} (originally from \cite{coco} cc-by-4.0), \cite{zerobench} (MIT) and \cite{robopatialhome} (apache-2.0)}
\label{fig:figure1}
\end{figure}

Several benchmarks have attempted to measure these shortcomings. However, many suffer from significant limitations: they often recycle existing datasets \cite{mmbench, pictureworth1000, spatial-mm, cosmosreason1, hsieh2023sugarcrepe, nuscenesspatial, sc++, vsr}, lack contrastive structure \cite{conme, robopatialhome, spatialbench, embspatialbench, blink}, or rely on synthetic or schematic images \cite{clevr, vismin, whatsupvlms} (see Figure \ref{fig:figure1}). As a result, these benchmarks tend to overestimate model performance, sometimes even allowing pure language models to score highly on supposedly vision-language tasks \cite{winoground, valse, balancedvqa}.

The reuse of older datasets raises concerns about data
contamination, as their contents may already be present in the training corpora of contemporary language models. Non-contrastive benchmarks often permit the use of unintended shortcuts, thereby failing to evaluate the specific capabilities they are designed to test. Furthermore, the inclusion of synthetic images in evaluation datasets poses challenges, as performance on such data does not reliably transfer to real-world scenarios. A notable example is the CLEVR dataset \cite{clevr}, where models have achieved near-saturating performance for years, despite continuing to struggle with real-image counterparts.

To address these issues, we introduce \textit{RocketScience}, a benchmark specifically designed to rigorously evaluate spatial understanding in VLMs. The dataset is comprised of 482 manually curated, contrastive image-text pairs representing diverse, real-world scenes (indoors, outdoors, across varying lighting conditions - see Appendix \ref{appendix:examples}). Each example forms a question-answer pair that is trivially solvable by humans within seconds, yet proves difficult for current vision-language models.

We evaluate three major categories of models: (1) dual-encoder models such as those in the CLIP family, (2) vanilla multimodal large language models (MLLMs), both open- and closed-source, and (3) advanced reasoning-based MLLMs like o4-mini and Gemini 2.5 Pro. Our results show that all models, except those explicitly designed for multimodal reasoning, perform at chance levels. In contrast, models utilizing chain-of-thought (CoT) prompting or reinforcement learning-based reasoning approaches approach near-perfect performance on this benchmark. We further disentangle model performance along two key axes: entity localization and spatial reasoning. Our analysis reveals that poor performance on spatial understanding tasks stems primarily from limitations in reasoning capabilities, rather than failures in localizing objects.

Our contributions are summarized as follows:
\begin{itemize}[leftmargin=*]
\item A new open-source, contrastive benchmark, RocketScience, built entirely from scratch using diverse, real-world (non-synthetic) data, specifically designed to evaluate spatial reasoning capabilities in VLMs.
\item An evaluation of three classes of models on the benchmark: CLIP-like models, VLMs and reasoning VLMs 
\item A disentangled analysis of reasoning-based model performance along two axes: object localization and spatial reasoning. We demonstrate that chain-of-thought reasoning is the primary bottleneck for solving spatial reasoning tasks.
\end{itemize}

\section{Related Work}
\label{sec:related-work}

\subsection{VLM Benchmarks and VLMs}
Several benchmarks have been proposed to evaluate vision language models in recent years. They span commonsense reasoning \cite{whoops, visualriddles}, understanding multiple images at the same time \cite{muirbench}, noticing small differences between images \cite{mismatchquest}, counting objects \cite{teachingcliptocount}, abductive reasoning \cite{nleye}, visual analogies \cite{kiva} and diagram understanding \cite{humanevalv}.
Larger benchmarks like OmniBench \cite{omnibench}, MMEvalPro~\cite{mmevalpro}, MMStar~\cite{rightwayevaluatingvlms} and WildVision \cite{wildvision} evaluate a wide range of VLM phenomena at the same time. 
Recently, particularly challenging benchmarks such as ZeroBench~\cite{zerobench} and Humanity's last exam~\cite{humanitysexam} have been released although it remains debatable whether a benchmark which is also hard for humans is even desirable, since the difficulty might be the result of poorly designed questions.

At the same time, vision-language models have also substantially improved in recent years. While traditional VLMs answer questions immediately, recently models have been trained to produce a chain-of-thought (CoT)~\cite{cot}, which is step-by-step reasoning, to improve their performance.

\subsection{Contrastive VLM Benchmarks}
A common problem with VLM benchmarks is that they can largely be solved using language models alone. \cite{valse, cappa, hsieh2023sugarcrepe, whatsleftcantberight, rightwayevaluatingvlms} This is because certain compositions of objects are more likely to appear in the world and questions about them can therefore be solved with common linguistic co-occurrences instead of visual understanding. To avoid this, contrastive benchmarks \cite{winoground, bivlc, vismin, rel3d, whatsupvlms, forest} are composed of so-called contrastive pairs, which are tuples of two images and two matching texts, that ideally only differ in the exact concept that we want to test for. This means that if the one caption within a contrastive pair contains a particularly likely scenario, like "a person on a horse" the other caption will have to be the unlikely opposite, "a person under a horse". Due to this contrastive design, a surface level statistical linguistic understanding is not enough to solve them and models need to have an integrated visio-linguistic understanding of the world to be able to solve them.

\subsection{Spatial Understanding}

Spatial understanding concerns reasoning about object positions and is evaluated across different modalities. For VLMs, benchmarks such as \cite{travlr, nlvr, visfactor, vlmsblind, deductivereasoning} test abstract spatial reasoning with simple shapes. Language-only datasets like SpartQA~\cite{spartqa} and WorldSense~\cite{worldsense} use question answering, while CVR~\cite{cvr} provides a visual outlier-detection task. Text-to-image benchmarks, including DrawBench~\cite{drawbench}, DALL-Eval~\cite{dall-eval}, and VISOR~\cite{benchspat}, assess spatial understanding through compositional prompts. Others, such as SpatialRGBT-Bench~\cite{spatialrgpt} and Q-Spatial Bench~\cite{reasoningpaths}, target distance reasoning. Among these diverse notions of spatial understanding, we focus specifically on spatial relations (e.g., on, under), as they form the foundation for more complex reasoning.

\subsection{Spatial Relation Benchmarks}

To measure progress on the understanding of spatial relations, a range of benchmarks have been developed over the past few years. We provide an extensive overview of them in Table~\ref{tab:relatedwork} in Appendix \ref{appendix:relatedwork} that also includes benchmarks with only a subset dedicated to spatial relations. A large portion of them is non-contrastive, making them potentially sensitive to shortcut solutions. Recycling data from other datasets instead of sourcing new data is common as well, which can lead to unrepresentative scores when this old data leaks into model training, inflating the scores. We will also show empirical evidence for those benchmarks being easier than RocketScience in a later section in Figure~\ref{fig:delta}. Additionally a few test sets like VisMin~\cite{vismin} are synthetically generated, which means we cannot directly infer models' competence on real data from their results, especially if the synthetic samples are very schematic. We know that slight changes applied to images can already change predictive models' outputs \cite{metposeestimation} \cite{imagetransforms}, so it is prudent not to assume that this domain transfer works automatically. Additionally, we can also still observe many artifacts and unrealistic appearing objects in synthetic data, although this is likely to improve with better image generation models in the future. Some of the older benchmarks are of significant size, which was common at the time, but can actually be expensive to evaluate on today, where most top models are not open source.

Winoground~\cite{winoground}, Naturalbench~\cite{naturalbench} and MMVP~\cite{mmvp-eyeswideshut} only contain a small number of spatial understanding questions and no identifiable subset evaluating this property specifically. Therefore they are not included in Table~\ref{tab:relatedwork}. CLEVR~\cite{clevr} is the oldest of the benchmarks. It contains synthetic images of abstract shapes and is very schematic.
VALSE~\cite{valse}, Visual Genome Relation ~\cite{vl-bow} and SpatialEval~\cite{pictureworth1000} are known to be solvable to a high degree by a blind language model without any image inputs.~\cite{valse, cappa, pictureworth1000} This likely also applies to many of the other non-contrastive benchmarks.
Some of the non-contrastive benchmarks like SugarCrepe~\cite{hsieh2023sugarcrepe}, ConMe \cite{conme} and SugarCrepe++ \cite{sc++} put a a great deal of effort into creating text foils (also known as \textit{hard negatives}) to choose from. NuScenes-SpatialQA \cite{nuscenesspatial} is a huge benchmark with automatically generated captions that only covers the self driving car domain.
Among the contrastive benchmarks, Rel3D~\cite{rel3d} is impressively diverse for a synthetic dataset.
FOREST~\cite{forest} and VSR~\cite{vsr} add perspective change from the perspective of objects, introducing additional complexity.

\section{RocketScience Benchmark}
\label{sec:methodology}

\begin{figure}[t]
\makebox[0pt][l]{%
  \adjustbox{trim=0pt 0pt 0pt 0pt,clip}{\includegraphics[width=0.098\textwidth]{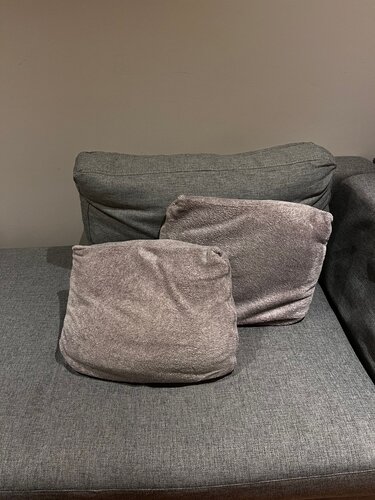}}%
  \adjustbox{trim=0pt 0pt 0pt 0pt,clip}{\includegraphics[width=0.098\textwidth]{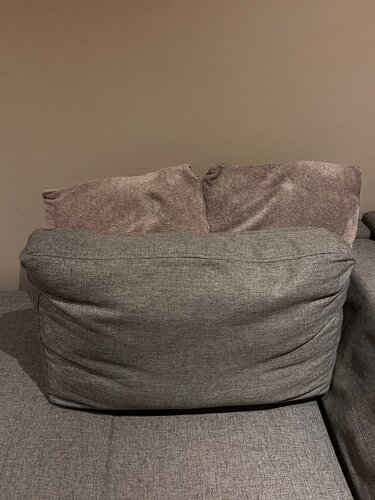}}%
  \adjustbox{trim=0pt 0pt 0pt 0pt,clip}{\includegraphics[width=0.098\textwidth]{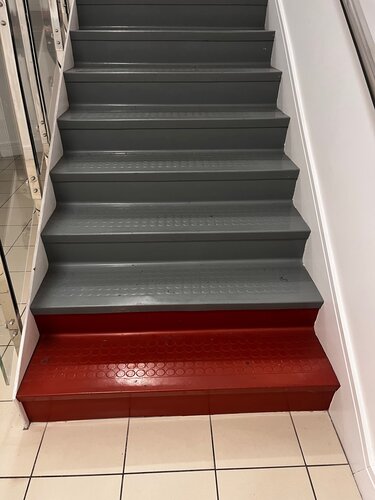}}%
  \adjustbox{trim=0pt 0pt 0pt 0pt,clip}{\includegraphics[width=0.098\textwidth]{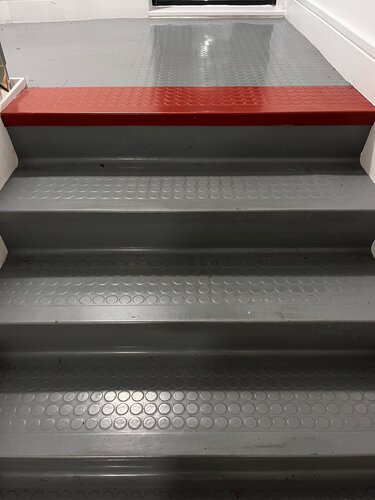}}%
  \adjustbox{trim=0pt 0pt 0pt 0pt,clip}{\includegraphics[width=0.098\textwidth]{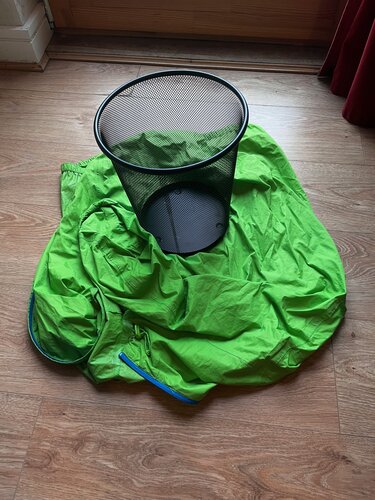}}%
  \adjustbox{trim=0pt 0pt 0pt 0pt,clip}{\includegraphics[width=0.098\textwidth]{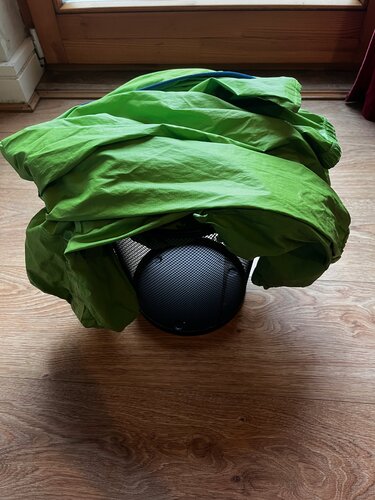}}%
  \adjustbox{trim=0pt 0pt 0pt 0pt,clip}{\includegraphics[width=0.098\textwidth]{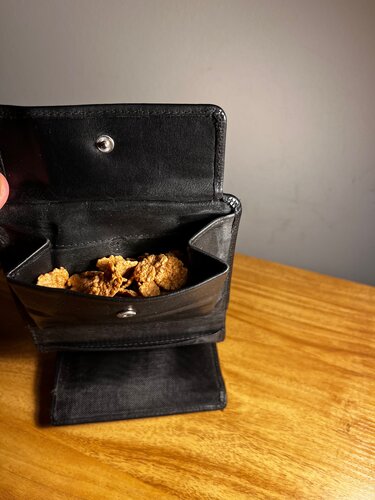}}%
  \adjustbox{trim=0pt 0pt 0pt 0pt,clip}{\includegraphics[width=0.098\textwidth]{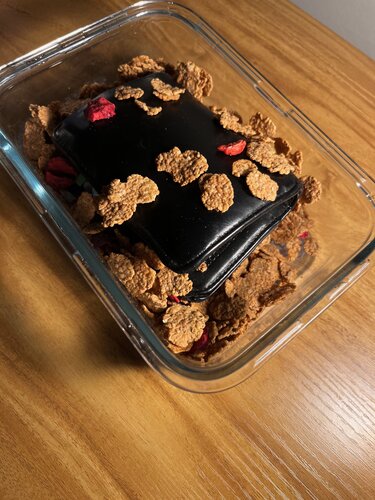}}%
  \adjustbox{trim=0pt 0pt 0pt 0pt,clip}{\includegraphics[width=0.098\textwidth]{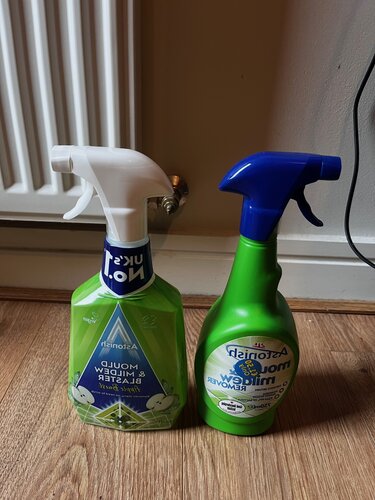}}%
  \adjustbox{trim=0pt 0pt 0pt 0pt,clip}{\includegraphics[width=0.098\textwidth]{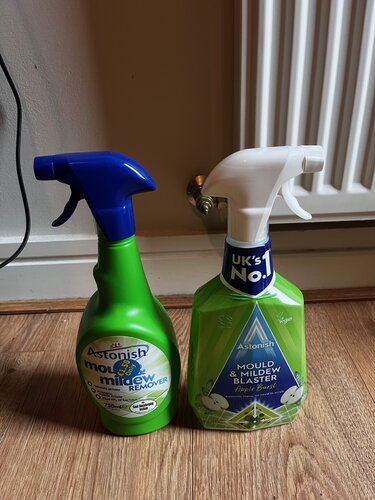}}%
}
\par\vspace{0pt}
\makebox[0pt][l]{%
  \adjustbox{trim=0pt 0pt 0pt 0pt,clip}{\includegraphics[width=0.098\textwidth]{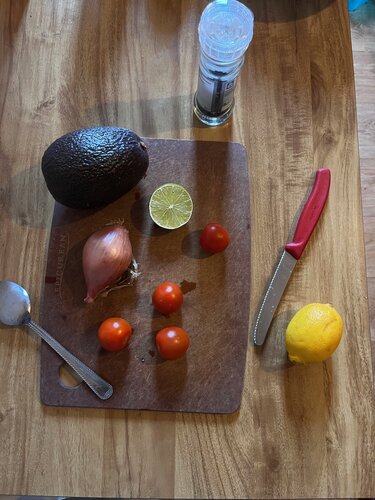}}%
  \adjustbox{trim=0pt 0pt 0pt 0pt,clip}{\includegraphics[width=0.098\textwidth]{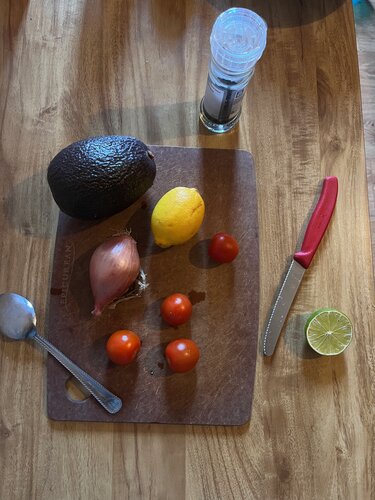}}%
  \adjustbox{trim=0pt 0pt 0pt 0pt,clip}{\includegraphics[width=0.098\textwidth]{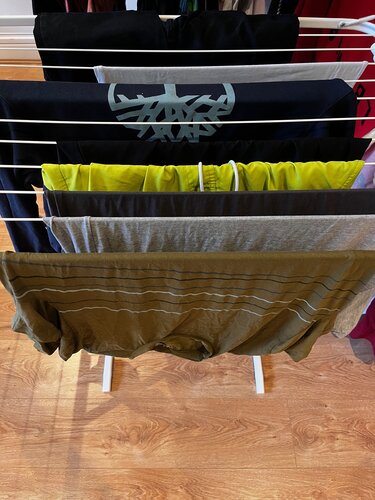}}%
  \adjustbox{trim=0pt 0pt 0pt 0pt,clip}{\includegraphics[width=0.098\textwidth]{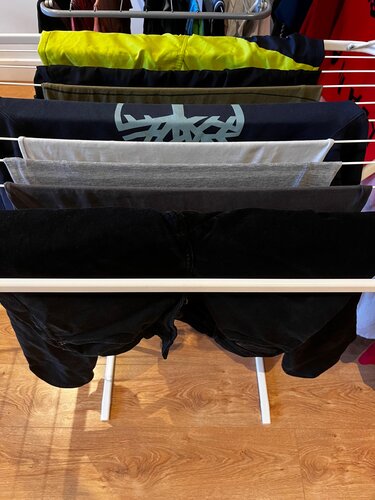}}%
  \adjustbox{trim=0pt 0pt 0pt 0pt,clip}{\includegraphics[width=0.098\textwidth]{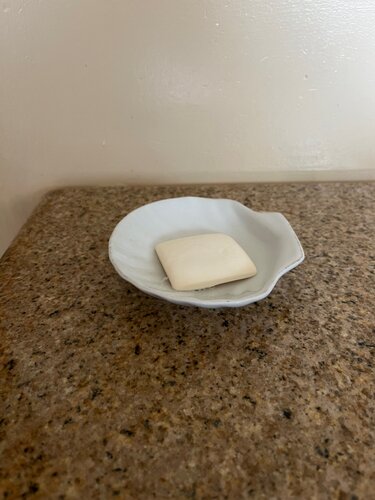}}%
  \adjustbox{trim=0pt 0pt 0pt 0pt,clip}{\includegraphics[width=0.098\textwidth]{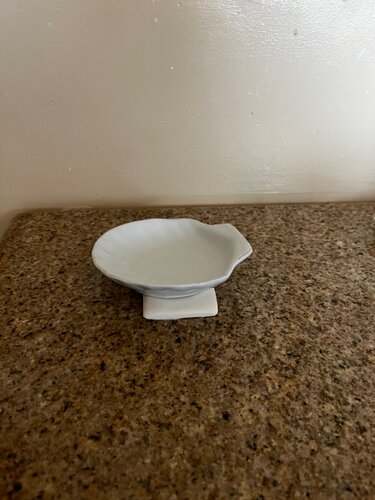}}%
  \adjustbox{trim=0pt 0pt 0pt 0pt,clip}{\includegraphics[width=0.098\textwidth]{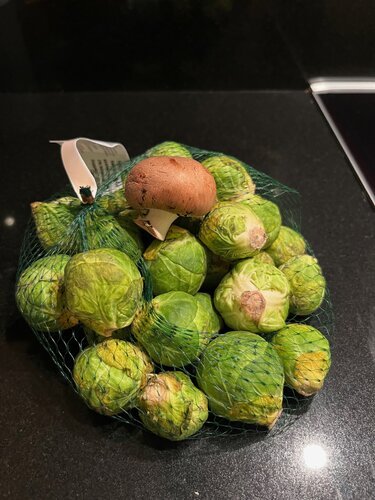}}%
  \adjustbox{trim=0pt 0pt 0pt 0pt,clip}{\includegraphics[width=0.098\textwidth]{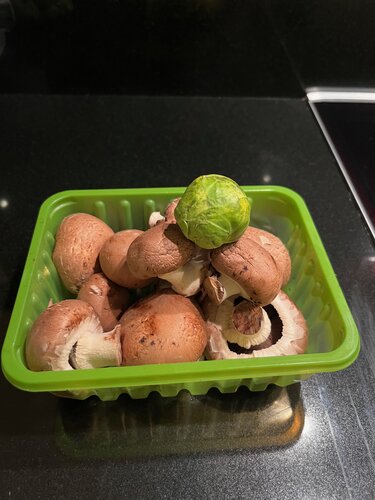}}%
  \adjustbox{trim=0pt 0pt 0pt 0pt,clip}{\includegraphics[width=0.098\textwidth]{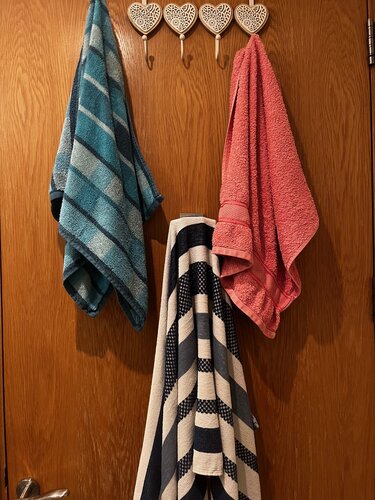}}%
  \adjustbox{trim=0pt 0pt 0pt 0pt,clip}{\includegraphics[width=0.098\textwidth]{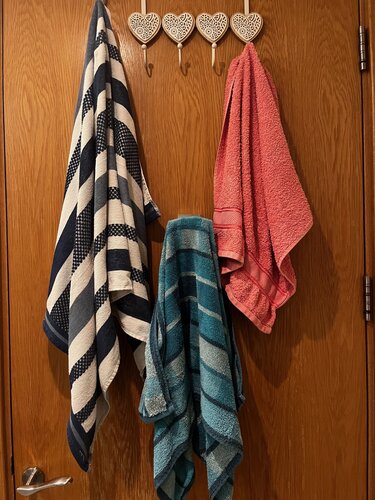}}%
}
\caption{Overview of the contrastive RocketScience dataset.}
    \label{examples}
\end{figure}

\subsection{Benchmark Design}

Based on the shortcomings outlined in Section~\ref{sec:related-work}, we have developed a carefully curated, real-world spatial understanding benchmark that includes a variety of different testing scenarios which avoid predefined schemas and that also follows a contrastive design to avoid shortcut solutions. (see Figure \ref{examples}) In contrast to other recent benchmarks like Zerobench~\cite{zerobench}, we create image-text pairs that are understandable to humans while being challenging for machines. We prove that examples are not challenging merely due to them being inherently ambiguous with extremely high human performance. (See Section \ref{section:humaneval})

The main focus of this benchmark is to systematically evaluate relative spatial understanding and the order of objects, as several works have suggested that these properties are not effectively learned by many vision-language models~\cite{hsieh2023sugarcrepe, whatsupvlms, whatsleftcantberight, vsr, spatial457}. We design the benchmark with contrastive pairs of new images and texts (each contrastive pair consists of two images and two matching captions with minimal differences). The benchmark requires two levels of understanding: object localization within the image and inference of their spatial relation. The questions are designed to require both steps, without shortcuts, by including the same objects in both parts of each contrastive pair.

\subsection{Data Collection}
We approximately balance the five main categories which all test for relative spatial understanding: \horizontal,  \vertical, \depth, \proximity and \order. The distribution of categories can be seen in Figure~\ref{fig:datasetdistribution}.

All images were collected in Europe and the USA with an iPhone 13 Mini. We intentionally exclude people and personally identifiable information for the purposes of data privacy. After collection, one author labeled the images and two additional authors checked their agreement with the labels, suggesting changes if necessary. We iterated this process until all authors agreed on the labels.

 We add \textit{Label} and \textit{Category} tags to each contrastive pair, where the label specifies the spatial relation necessary to distinguish the two samples in the pair (e.g. "left of") and the category indicates the spatial category (e.g. \horizontal). A contrastive pair can have two categories when the spatial relations are not polar opposites (e.g. "shoe in front of box" vs. "shoe to the right of box" would be assigned the categories \horizontal and \depth).

\subsubsection{Images}
\label{sec:data_collection:images}
As some objects are impossible or extremely challenging to move in the real world, many of the pairs contrasting left and right positions consist of an original photo and its mirrored version as its opposite. Mirroring was only applied when no relevant text would be distorted through of mirroring.

To account for variation in lighting, images were captured in different lighting scenarios. These included: natural light outside, natural light inside, nighttime outside and artificial light inside. We note, however, that the lighting conditions are always consistent within a single contrastive pair.

\subsubsection{Captions}
Captions were designed such that the two captions for a pair of images could only differ in the position of the objects. For example:
\begin{itemize}[leftmargin=*]
  \item \textbf{Word order}: This represents the largest share of the dataset, e.g. "A chair to the left of a table" vs. "A table to the left of a chair"
  \item \textbf{Swapped Preposition}: This is necessary for some relations that cannot be inverted by simply changing the word order, e.g. "The scissors are close to the door" vs. "The scissors are far from the door".
\end{itemize}
In both cases, the semantics of the two captions are opposites (also termed \textit{hard negatives}) with respect to the objects' positions. These hard negatives are required to assess whether a spatial relation has been successfully learned rather than simply the likelihood of a relation being associated with certain nouns (i.e. people on chairs, not chairs on people). With the exception of the "left" vs. "right" distinction, in many cases, hard negatives control for unlikely noun-relation cases, while a benchmark without hard negatives simply checks whether models have memorized the most likely case.

\subsection{Scope and Variety}

The benchmark is designed for spatial understanding, but the localization of the objects still requires additional understanding such as: counting, negation, quantifiers ("most of"), materials, size and colour. 
In comparison to What's Up \cite{whatsupvlms}, RocketScience is significantly more diverse and less schematic. We include a wide variety of scene characteristics  such as indoor and outdoor environments, daytime and nighttime settings, objects at
varying distances (both near and far), a range of object sizes (small and large), as well as natural and rural environments. (See Appendix \ref{appendix:examples}) This broad coverage makes our dataset substantially more
representative of real-world conditions. 

The physical objects found in the images are sourced from a wide range of objects from daily life within Europe and the US. The distribution of objects can be seen in Figure~\ref{fig:physicalobjectcounts} in Appendix \ref{appendix:datasetanalysis}. The most frequent adjectives are colors, but materials like \textit{"wooden"} or \textit{"metal"}, in addition to counts and other object properties can be found in the dataset, see~ Figure~\ref{fig:adjectivetcounts} which is also in Appendix \ref{appendix:datasetanalysis}.

\subsection{Ambiguity}

By design, most of the questions compare direct opposites, making ambiguity rare, however samples involving relations that are not direct opposites leave more room for interpretation. To prove solvability and to ensure unambiguous design we measure human performance. The humans obtain extremely high scores with low variance, from which we conclude that it is not ambiguous. (See Section \ref{section:humaneval} for details)

To ensure clarity in spatial references, we adopt the camera view as the default perspective for all annotations. Since the dataset excludes humans, we eliminate potential ambiguity arising from subjective references such as “on a person's left/right side”. The camera-based viewpoint provides a consistent and semantically grounded interface between visual and linguistic modalities, reducing interpretational variability.
Although most prompts contrast clearly defined opposites minimizing ambiguity, certain spatial relations, such as “near” vs. “far”, or comparative attributes like “big” vs. “small”, can be inherently subjective. To mitigate this subjectivity, each instance presents the model with two alternatives (e.g. a pair of captions or images), ensuring that the comparative context disambiguates the intended meaning. For instance, given two images and the caption “scissors close to the door”, the correct answer becomes the one where the scissors are closer to the door relative to the alternative.
As discussed in Section~\ref{sec:data_collection:images}, a subset of examples in the \horizontal category includes horizontally mirrored images. In some cases, mirrored text may be visible, but never in a way that would simplify the task.

\subsection{Difficulty}
Even in contrastive benchmarks, considerable care must be taken to prevent models from exploiting unintended correlations or shortcut signals (i.e. solving tasks via surface-level cues rather than the intended reasoning). In our benchmark, samples are designed to control for this by ensuring that both target entities are present in each image. For example, in the contrastive pair “beetroot in a bin” vs. “no beetroot in a bin”, both the beetroot and the bin appear in both images. The latter image still contains the beetroot, but located outside the bin. This ensures that solving the task requires understanding the spatial relation, rather than simply detecting object presence. This construction prevents the task from being reduced to simple object detection and enforces a requirement for spatial comprehension.

Additionally, a subset of image-text pairs is constructed to require fine-grained attribute localization, rather than reliance on nouns alone. For instance, in a caption such as “a crushed milk carton to the left of an intact milk carton”, both objects belong to the same category, but differ in visual attributes. Tasks of this nature increase linguistic complexity and demand comprehension of adjectives in context, further discouraging reliance on shallow pattern matching.

\begin{figure}[t]
  \centering
  \begin{subfigure}[b]{0.65\textwidth}
    \centering
    \includegraphics[trim=0 0 0 100, clip,width=\linewidth]{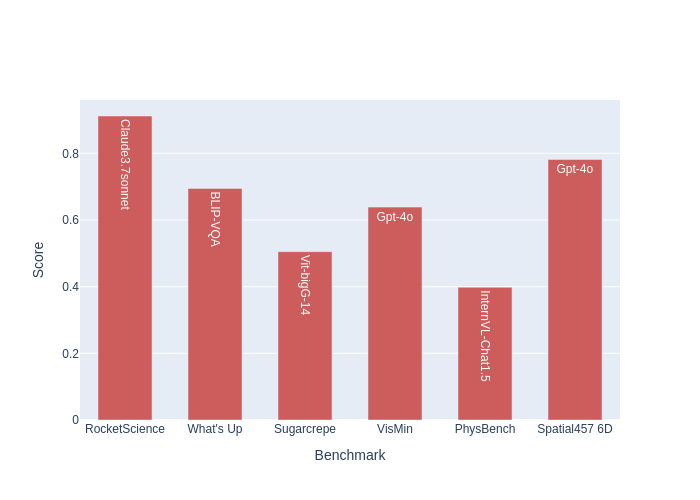}  
    \caption{}
    \label{fig:delta}
    \end{subfigure}
  \hfill
  \begin{subfigure}[b]{0.33\textwidth}
    \centering
    \raisebox{8mm}{\includegraphics[width=\linewidth]{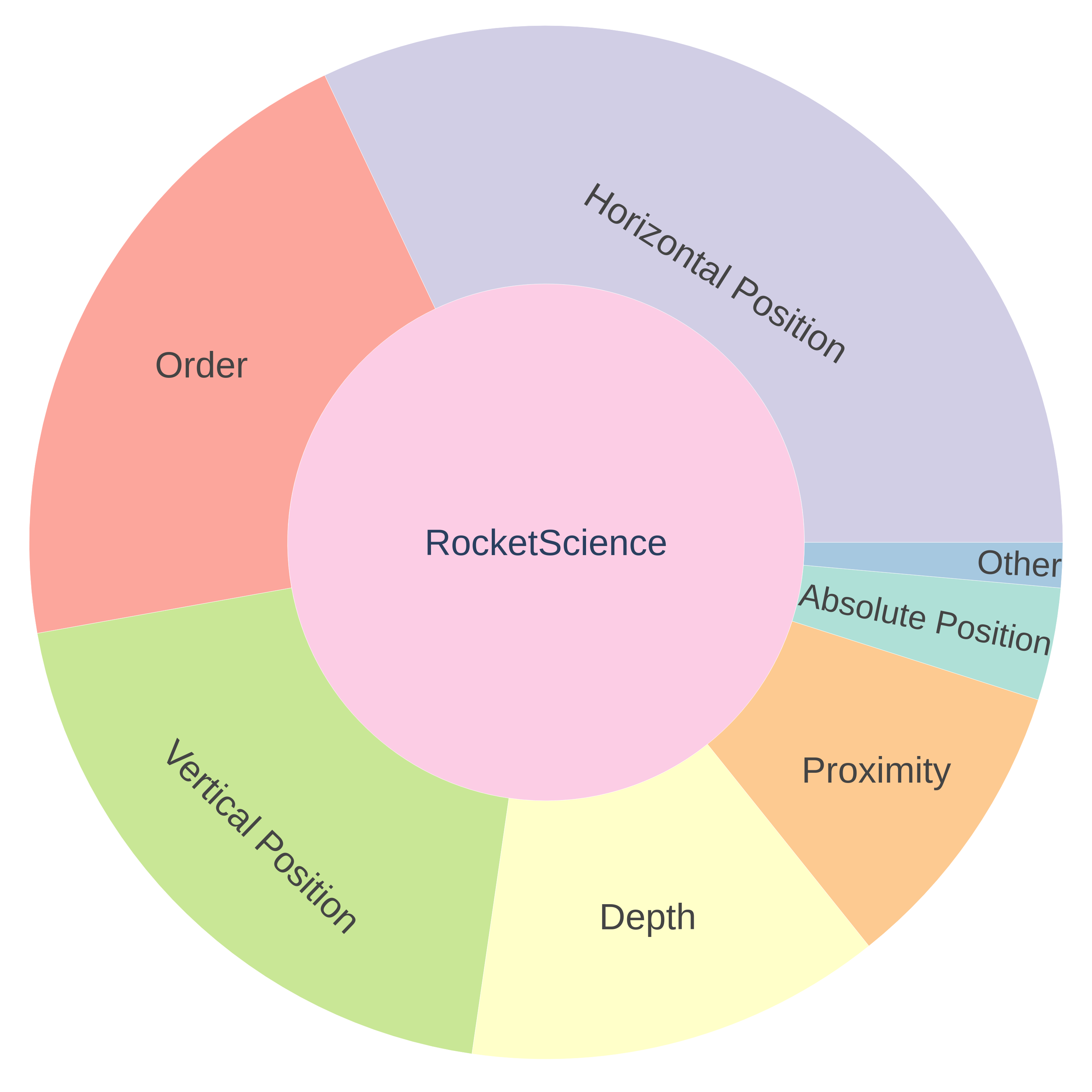}}
    \caption{}
    \label{fig:subfig2}
  \end{subfigure}

  \caption{Overview of the RocketScience benchmark. 
\textbf{(a)} Benchmark quality score for popular benchmarks where higher values indicate greater challenge and headroom. 
\textbf{(b)} Categories contained in the RocketScience dataset. (Full breakdown in Figure \ref{fig:datasetdistribution} in Appendix \ref{appendix:datasetanalysis})}
  \label{fig:mainfig}
\end{figure}

To account for the fact that benchmark datasets often leak into model training over time, we argue that benchmark difficulty and informativeness are best evaluated at the time of their release. To quantify this, we introduce a normalized scoring metric:
\begin{equation}
    \text{Score} = \frac{1 - \text{SOTA}_{\text{AtRelease}}}{1 - \text{Random}}
\end{equation}
 
where 
$\text{SOTA}_{\text{AtRelease}}$
  is the best-performing \textit{non-CoT} model at the time of dataset release, and 
$\text{Random}$ represents the expected score from uniform guessing. This metric captures the benchmark's potential to differentiate between random and competent models at release time, rather than its current saturation level. It also allows for fairer comparisons across datasets of varying ages. Under this metric, our benchmark RocketScience outperforms several contemporaneous datasets as can be observed in figure \ref{fig:delta}, even though the metric favors older benchmarks as they do not have to compare to current SOTA models.

\section{Experimental Setup \& Evaluation}
\label{sec:experimental-setup}

\subsection{Models}
We evaluate three main types of models: CLIP-like (contrastive) embedding models, language models with vision capabilties (VLMs) and reasoning VLMs with a hidden chain-of-thought (reflective models). The most interesting models in our context are those that have been trained for more fine-grained visual understanding like NegCLIP \cite{vl-bow}, Paligemma \cite{paligemma}, SpaceOm \cite{spaceOm} and GLM4.1 \cite{glm41thinking}. We also select a range of VLMs from top frontier VLMs like GPT-4o \cite{gpt4o} and Llama4 \cite{llama4}. Additionally, we also evaluate Gemini 2.5 \cite{gemini2.5} and o4-mini \cite{o4-mini} as examples of reflective models.
We only include commercial models that can be run for less than ten US Dollars on the benchmark and only run them once due to their cost.

\subsection{Evaluation}
\textbf{Preprocessing}: We resize all images to $1024\times1024$. Most of the CLIP models and Paligemma include custom preprocessing that downsizes images further. Closed VLMs like GPT-4o and Claude Sonnet allow for large image inputs. For all API models we use \ensuremath{\mathtt{.png}} to maintain image quality.

\textbf{Inputs}: Each contrastive pair is split into four questions:

Q1: First image + both captions $\rightarrow$ Which is the correct caption?\\
Q2: Second image + both captions $\rightarrow$ Which is the correct caption?\\
Q3: First text + both images $\rightarrow$ Which is the correct image?\\
Q4: Second text + both images $\rightarrow$ Which is the correct image?

The correct answer for Q1 and Q3 is always 1 and for Q2 and Q4 it‘s always 2. This automatic alternation means it is not necessary to shuffle the answers as is necessary for non-contrastive benchmarks to avoid models simply always choosing the first answer.

CLIP-like models receive the images and captions individually and compute the similarity between them. The exact inputs and outputs can be seen in Appendix \ref{appendix:modeldetails}.

\textbf{Decoding}: For API models we set temperature = 0 where possible, to minimize variance. Please note that this still does not lead to fully deterministic according to their documentation \cite{gpt4o, gemini2.5, llama4}. CLIP models are deterministic during evaluation. For local VLMs, we use greedy decoding in order to ensure results are reproducible.

\textbf{Hardware and Runtime}:
The local models were evaluated on a T4-GPU. Most CLIP models evaluate within a few minutes. The API models take between 1 and 2.5 hours to run, with the reflective models taking the most time.

\subsection{Metrics}
\label{section:metrics}
We use text score, image score and group score, as introduced by Winoground~\cite{winoground}, for the CLIP-like models and modify the scores to be suitable for VLMs as follows: for text score, one image and two captions are used as input with the model then choosing the matching caption. For image score, two images and one caption act as the input and the model must select the correct image. This approach is not perfectly comparable to the original scores, as they are based on individually encoded texts and images, making them slightly harder than our newly defined VLM scores. The adapted image score, text score and group score are defined in Equations~\ref{eq:image_score}, \ref{eq:text_score} and \ref{eq:group_score} as follows: 
\begin{equation}
\label{eq:image_score}
f(I_0, I_1, T_0, T_1) =
\begin{cases}
1, & \text{if } \hat{y}(I_0, I_1, T_0) = \text{``choose } I_0 \text{''} \text{ and } \hat{y}(I_0, I_1, T_1) = \text{``choose } I_1 \text{''} \\
0, & \text{otherwise}
\end{cases}
\end{equation}

\begin{equation}
\label{eq:text_score}
g(I_0, I_1, T_0, T_1) =
\begin{cases}
1, & \text{if } \hat{y}(I_0, T_0, T_1) = \text{``choose } T_0 \text{''} \text{ and } \hat{y}(I_1, T_0, T_1) = \text{``choose } T_1 \text{''} \\
0, & \text{otherwise}
\end{cases}
\end{equation}

\begin{equation}
\label{eq:group_score}
h(I_0, I_1, T_0, T_1) =
\begin{cases}
1, & \text{if } f(I_0, I_1, T_0, T_1) = 1 \text{ and } g(I_0, I_1, T_0, T_1) = 1 \\
0, & \text{otherwise}
\end{cases}
\end{equation}

where $f$, $g$, $h$ are the equivalents of image score, text score and group score and $I_0, I_1, T_0, T_1$ are the images and texts from a contrastive pair.

\subsection{Human Evaluation}
\label{section:humaneval}
To assess the solvability of the benchmark we present the questions to humans with no formal linguistics education. (n=4) The evaluation scheme is the same as for VLMs, so they either get a caption and have to choose between two images or get an image and have to choose the right of two captions. However because humans can remember their previous responses, we only present one question from each contrastive pair to each human. The scores are then computed across all testers. For humans the two images or the two captions are shuffled - otherwise the correct one would always be in the same spot. The image resolution for the human evaluation is 600 by 600 pixels so that two images fit onto a laptop screen comfortably. (This is lower than the resolution the models receive) The exact instructions and interface are available in Appendix \ref{appendix:humanbaseline}.
Our participants score a mean accuracy of $0.985$ with a standard deviation of $0.008$. We conclude that RocketScience has an extremely low level of ambiguity.

\section{Results}
\label{sec:results}

We present the results in Table \ref{smallresults}. Table \ref{table:bigresults} in Appendix \ref{appendix:detailedresults} also includes text scores and image scores. It also contains additional CLIP-like models. Model names that end with \textit{\_cot} have been prompted to do chain-of-thought reasoning on top of their usual system prompt. Gemini 2.5 and o4-mini perform reasoning internally before responding. We find that all open vision-language models, even those trained for spatial understanding perform very poorly and often below random chance. This phenomenon has also been observed in other contrastive vision-language benchmarks, such as Winoground \cite{winoground} and WhatsUp \cite{whatsupvlms}. Only reasoning models come close to human performance.

\begin{table}[h]
\centering
\begin{tabular}{lrrrrrrr}
\toprule
Model Name & Horiz & Vert  & Depth & Prox & Order & Abs Pos & Total \\
\midrule
Random chance & 0.17 & 0.17 & 0.17 & 0.17 & 0.17 & 0.17 & 0.17 \\
Human & 0.96 & 0.98 & 0.95 & 1.00 & 0.92 & 0.80 & 0.95\\
\midrule
ViT-B-32negCLIP \cite{vl-bow} & 0.00 & 0.02 & 0.00 &  0.00 & 0.03 & 0.00 & 0.01 \\
CoCa\_ViT-L \cite{coca} & 0.00 &  0.04 & 0.00 & 0.00 & 0.02 & 0.00 & 0.01 \\
PaliGemma-3b-mix-448 \cite{paligemma} & 0.01 & 0.04 & 0.00 & 0.00 & 0.02 & 0.20 & 0.02 \\
Qwen-2.5-vl-72b-instruct \cite{qwen2.5vl} & 0.00 & 0.00 & 0.00 & 0.00 & 0.00 & 0.00 & 0.00 \\
Qwen-vl-max \cite{qwenvlmax}& 0.01 & 0.02 & 0.00 & 0.00 & 0.02 & 0.00 & 0.01 \\
Claude-3-7-sonnet-20250219 \cite{claude3.7}& 0.17 & 0.38 & 0.11 & 0.52 & 0.18 & 0.00 & 0.24 \\
Llama-4-maverick \cite{llama4}& 0.17 & 0.23 & 0.05 & 0.52 & 0.15 & 0.40 & 0.20 \\
GPT-4o-2024-08-06 \cite{gpt4o}& 0.08 & 0.40 & 0.24 & 0.40 & 0.07 & 0.00 & 0.19 \\
\hlorange{Llama-4-maverick\_cot} \cite{llama4}& 0.53 & 0.42 & 0.21 & 0.64 & 0.45 & 0.20 & 0.44 \\
\hlorange{GPT-4o-2024-08-06\_cot} \cite{gpt4o}& 0.44 & 0.68 & 0.39 & 0.56 & 0.52 & 0.60 & 0.51 \\
\hlorange{SpaceOm} \cite{spaceOm}& 0.01 & 0.02 & 0.03 & 0.04 & 0.00 & 0.00 & 0.01 \\
\hlgreen{Glm-4.1v-9b-thinking} \cite{glm41thinking} & 0.77 & 0.68 & 0.42 & 0.4 & 0.68 & 1.00 & 0.64 \\
\hlgreen{Gemini-2.5-pro-preview-03-25 }\cite{gemini2.5}& 0.91 & 0.87 & 0.76 & \textbf{0.76} & 0.80 & 0.80 & 0.83 \\
\hlgreen{o4-mini} \cite{o4-mini}& \textbf{0.97} & \textbf{0.91} & \textbf{0.89} & 0.68 & \textbf{0.88} & \textbf{1.00} & \textbf{0.89} \\
\bottomrule
\end{tabular}
\caption{Group score on RocketScience by category of the task (Horizontal Position, Vertical Position, Depth, Proximity, Order, Absolute Position, Total Score). The highest score in each category is bold. Models run with chain-of-thought prompting are marked in \hlorange{orange} and reflective models are marked in \hlgreen{green}.}
\label{smallresults}
\end{table}

\subsection{Why Are Chain-of-Thought Models Better?}
\begin{figure}[t]
  \centering
  \begin{subfigure}[b]{0.40\textwidth}
    \centering
    \begin{tabular}{lcc}
      \toprule
      Model & Acc lf & Acc rf \\
      \midrule
      gpt-4o & $96.11\pm0.96 $& $90.55\pm0.96$ \\
      o4-mini & $96.66\pm 0.00$ & $95.56\pm1.92$  \\
      \bottomrule
    \end{tabular}
    \caption{Mean accuracy and standard deviation over 3 runs on the localization task (horizontal position subset). The difference in localization performance between non-CoT and CoT models is minimal. A small performance gap is also observed based on object order in the prompt: accuracy is slightly higher when the first-mentioned object appears on the left (Acc lf) compared to when it appears on the right (Acc rf), for both model types.}
    \label{fig:localisation}
  \end{subfigure}
  \hfill
  \begin{subfigure}[b]{0.58\textwidth}
    \centering
    \includegraphics[trim=0 40 50 100, clip, width=\linewidth]{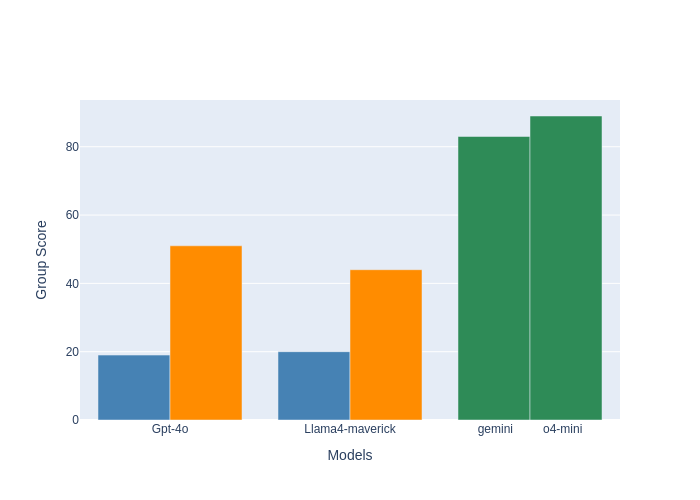}
    \caption{Group score comparison of models without chain-of-thought (blue), with explicit chain-of-thought prompting (orange), and with implicit chain-of-thought reasoning (green).}
    \label{fig:cot}
  \end{subfigure}

  \caption{Disentanglement experiments of (a) localisation and (b) CoT reasoning}
  \label{fig:cotresults}
\end{figure}

We hypothesize that there are two main steps necessary to detect a spatial relation: localization of objects and inference of spatial relation. We examine both stages to determine why reasoning models like o4-mini perform much better than their non-reasoning counterparts like gpt-4o:
\begin{itemize}[leftmargin=*]
    \item \textbf{Localization of objects}: We test whether o4-mini is better at localizing the objects than gpt-4o. We use the horizontal position subset for this (on which gpt-4o performs very poorly). The two models are prompted to provide bounding boxes for both objects in each image. We then check whether the coordinates of both objects are in the correct spatial configuration. (not exact location) Figure \ref{fig:localisation} shows the results. We find that gpt-4o is very close to o4-mini's performance, indicating that reasoning does not help with the localisation stage, but only with concluding the correct spatial relation.
    \item \textbf{Inference of spatial relation}: We use two sets of prompts for non-reflective models: The first prompts each model to output the answer alone immediately and nothing else (non-CoT). The second prompts each model to first reason and then output the results (CoT). We find that the second prompt significantly improves performance over the first one, indicating that chain-of-thought reasoning plays the biggest role in the improved performance. (See Figure \ref{fig:cot})
\end{itemize}

\section{Discussion}
\label{sec:conclusion}
\subsection{Limitations}

First, models accessed via APIs were only evaluated once due to cost constraints. Although we set the sampling temperature to zero where possible to ensure determinism, outputs may still vary slightly between runs. Second, while our dataset aims to reflect real-world complexity, some scenes remain less cluttered with objects than typical real-world environments. Introducing additional clutter while maintaining clear, unambiguous relations remains a significant challenge. Third, we observe notable performance improvements using chain-of-thought prompting; however, this benefit may be specific to top-tier commercial models and might not generalize to smaller or open-source models. Finally, although we strive to minimize changes within each contrastive pair, slight variations in camera angle may occur. This could introduce a potential shortcut where models exploit angle differences to infer spatial relations rather than relying solely on object configurations.

\subsection{Ethics}

People and personal data are explicitly excluded as subjects within this dataset in an effort to minimize the unnecessary use of human data in experiments. Each sample within the dataset was meticulously reviewed by three of the authors of this paper in an effort to maintain quality standards by minimizing errors and omissions. It is necessary to emphasize that the geographical locations of images are limited to the US and Europe, making the environment and objects specific to these locations. It is therefore crucial to stress that a model that performs well on this benchmark will not necessarily perform well in all geographic locations or with objects specific to them. Additionally, though this benchmark has successfully revealed that most vision-language models fail to effectively model spatial relations, it should not be relied upon alone as a form of evaluation for these relations, as benchmark performance should not act as a replacement for application and location-specific quality control and testing. We caution that though benchmarks can reveal \textit{some} model shortcomings, they are never exhaustive and we caution against their use alone to rank algorithms in real-world applications/production as this can lead to unexpected societal outcomes.

\subsection{Conclusion}
We introduce RocketScience, a challenging new benchmark for evaluating spatial relation understanding in vision language models. 
Built from scratch using real-world, contrastive image-text pairs, RocketScience reveals that most open-source and commercial models perform surprisingly poorly. Our analysis shows that chain-of-thought prompting significantly improves performance. By disentangling the effects of object localization and spatial reasoning, we find that the primary limitation lies in models' ability to perform structured reasoning about spatial relations, rather than in their visual perception. We hope RocketScience serves as a diagnostic and development tool for future VLMs, encouraging research toward models with more robust spatial understanding.

\section*{Acknowledgements and Funding}
This publication has emanated from research conducted with the financial support of Science Foundation Ireland under Grant number 18/CRT/6183. For the purpose of Open Access, the author has applied a CC BY public copyright licence to any Author Accepted Manuscript version arising from this submission.

\bibliography{bibliography}

\bibliographystyle{plainnat}


\appendix

\clearpage
\section{Appendix: Related Work}
\label{appendix:relatedwork}

\begin{table}[h]
\centering
\begin{tabular}{lccccl}
\toprule
Benchmark & contrastive & new data & real data & non-schematic & size \\
\midrule

MMBench\cite{mmbench} & \xmark & \xmark & \cmark & \cmark & 125 \\
SpatialEval-Real \cite{pictureworth1000} & \xmark & \xmark & \cmark & \cmark & 1000 \\
VSR \cite{vsr} & \xmark & \xmark & \cmark & \cmark & 2195 \\
CLEVR \cite{clevr} & \xmark & \cmark & \xmark & \xmark & 15,000  \\
VALSE \cite{valse}& \xmark & \xmark & \cmark & \cmark & 535  \\
SugarCrepe \cite{hsieh2023sugarcrepe}& \xmark & \xmark & \cmark & \cmark & 1406  \\
ConMe \cite{conme} & \xmark & \xmark & \cmark & \cmark &  6793\\
SC++ \cite{sc++} & \xmark & \xmark & \cmark & \cmark & 1406 \\
VGR (ARO) \cite{vl-bow}& \xmark & \xmark & \cmark & \cmark & 23,937\\
RoboSpatial-Home \cite{robopatialhome}& \xmark & \cmark & \cmark & \cmark &123  \\
BLINK\cite{blink} & \xmark & \xmark & \cmark & \cmark &286  \\
SpatialBench \cite{spatialbench}& \xmark & \cmark & \cmark & \cmark & 35  \\
Space3D-bench \cite{space3dbench}& \xmark & \xmark & \cmark & \cmark & 188  \\
Spatial-MM \cite{spatial-mm}& \xmark & \xmark & \cmark & \cmark & 2,000  \\
EmbSpatial Bench \cite{embspatialbench} & \xmark & \xmark & \cmark & \cmark & 3,640  \\
NuScenes-SpatialQA \cite{nuscenesspatial} & \xmark & \xmark & \cmark & \cmark & 2,500,000 \\
Cosmos1 \cite{cosmosreason1}& \xmark & \xmark & \cmark & \cmark & 292  \\
Spatial457 \cite{spatial457}& \xmark & \cmark & \xmark & \cmark & 9,990 \\
FOREST \cite{forest} & \cmark & \cmark & \xmark & \cmark & 4,352 \\
BiVLC \cite{bivlc}& \cmark & \cmark & \xmark & \cmark & 1,400  \\
Rel3D \cite{rel3d} & \cmark & \cmark & \xmark & \cmark & 27,336 \\
VisMin \cite{vismin} & \cmark & \cmark & \xmark & \xmark & 622 \\
What's Up (A+B) \cite{whatsupvlms}& \cmark & \cmark & \cmark & \xmark & 820  \\
\textbf{RocketScience} & \cmark & \cmark & \cmark & \cmark & 482 \\

\bottomrule
\end{tabular}
\caption{Overview of Spatial Relations benchmarks (or with a subset for that). Contrastive means fully contrastive, new data means entirely new and no recycled parts, real data means not synthetic, non-schematic means different scenes and objects not always in the same positions, size is only the spatial relations subset and measured as number of image-text pairs.}
\label{tab:relatedwork}
\end{table}

\clearpage
\section{Appendix: Inputs and Outputs}
\label{appendix:modeldetails}
\begin{figure}[h!]  
    \centering
    \includegraphics[width=0.8\textwidth]{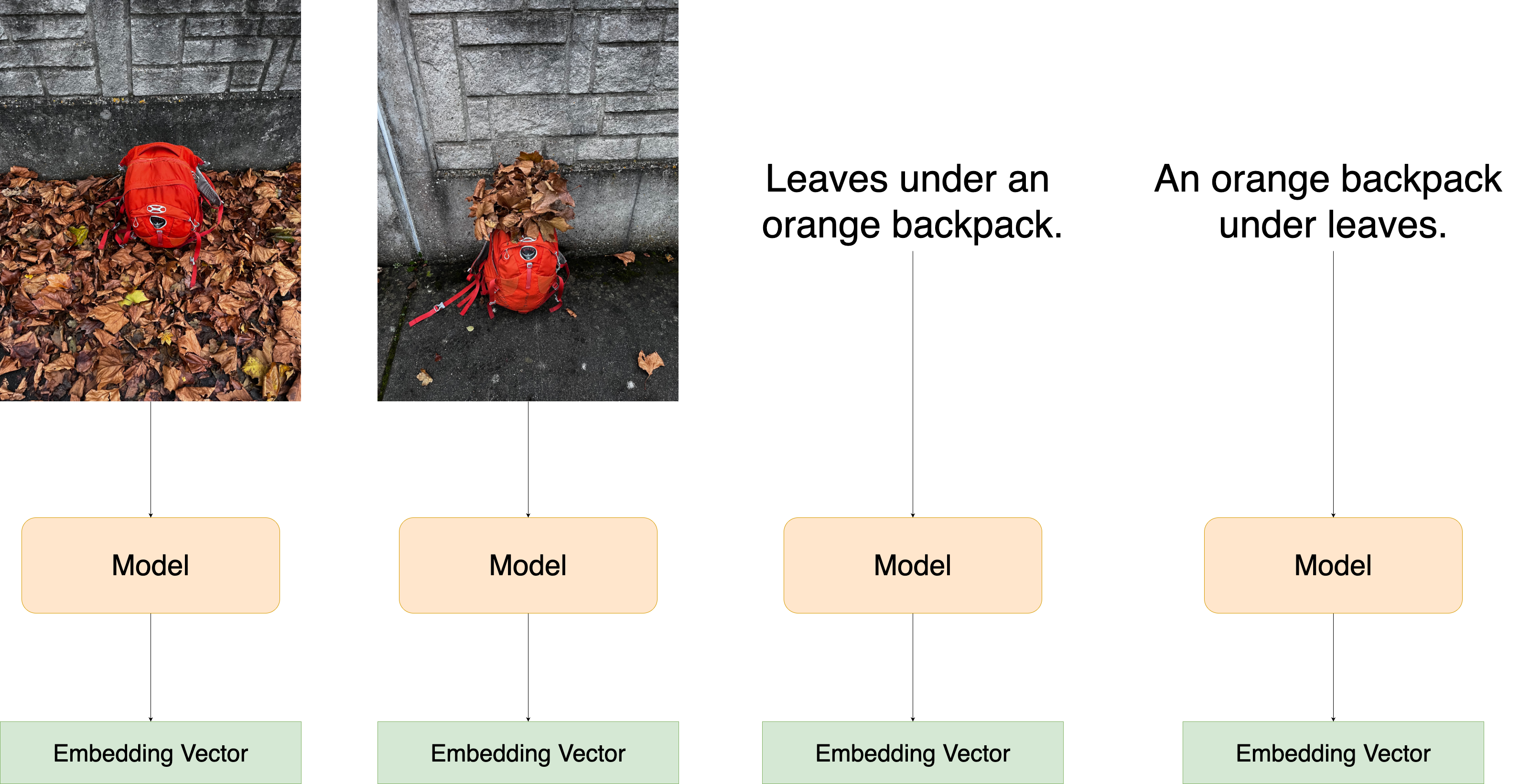}
    \caption{Function of CLIP-like models: they embed each image and text into a vector independently without having access to the other inputs at the same time}
    \label{appendix:CLIP-function}
\end{figure}

\begin{figure}[h!]  
    \centering
    \includegraphics[width=0.8\textwidth]{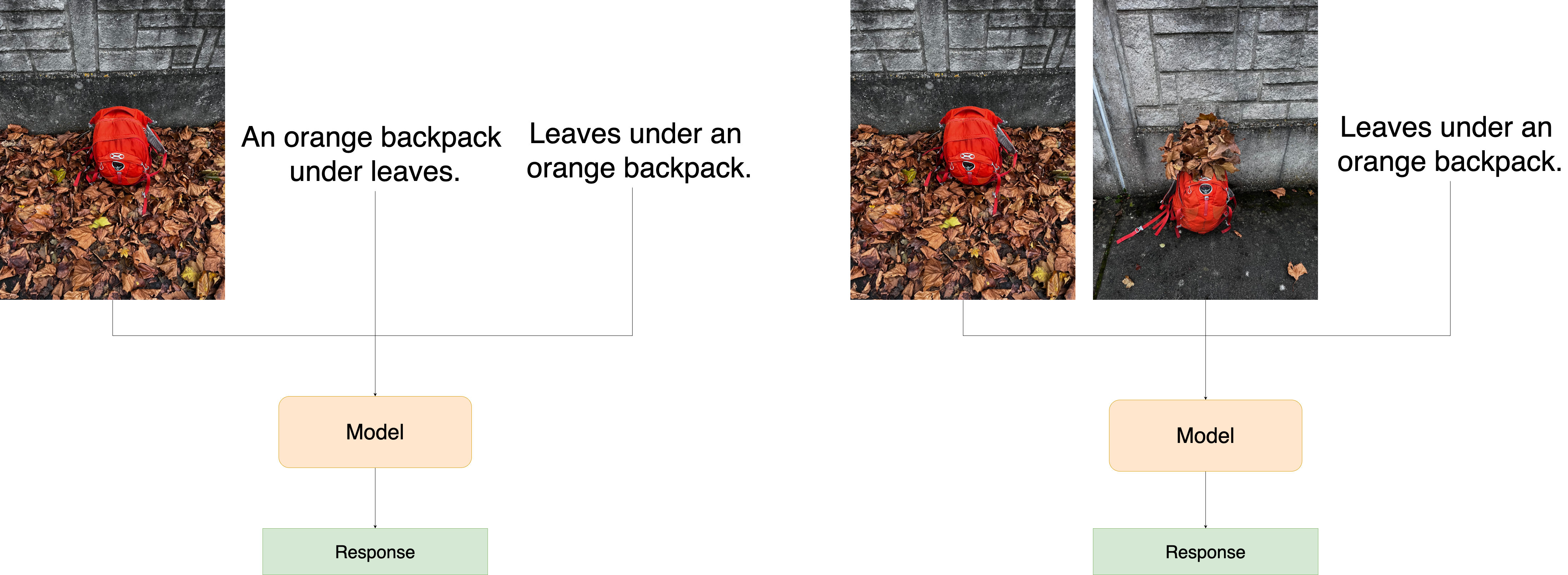}
    \caption{Function of VLMs: they receive either an image and two captions and have to choose a caption or they receive two images and a caption and have to choose an image  }
    \label{appendix:VLM-function}
\end{figure}

The model prompts are as follows:
CLIP models simply receive the captions from the dataset without additional prompting. The other VLMs (both normal and reflective) receive two types of prompts:
\begin{itemize}
    \item "Which caption fits the image best? Reply only with the number 1 or 2, nothing else. 1.) [CAPTION1] 2.) [CAPTION2]"
    \item "Which image fits the caption best? Reply only with the number 1 or 2, nothing else. Caption: [CAPTION1]"
\end{itemize}
For our additional experiments we test models' explicit chain-of-thought capabilities. These models are denoted with \_cot after their name. THey receive different prompts to make them reason:
\begin{itemize}
    \item "Which caption fits the image best? Reason about it and at the end write RESPONSE and reply only with the number 1 or 2. 1.) [CAPTION1] 2.) [CAPTION2]"
    \item "Which image fits the caption best? Reason about it and at the end write RESPONSE and reply only with the number 1 or 2. Caption: [CAPTION1]"
\end{itemize}

Example reasoning trace from gpt-4o for a case with two images and one caption: The caption describes "A grey bin on a white towel." Image 1 shows a grey bin placed directly on a white towel. Image 2 shows a towel covering the grey bin.The best fit for the caption is image 1, as it correctly shows the bin on the towel. RESPONSE 1

\clearpage
\section{Appendix: Selected examples from the dataset}
\label{appendix:examples}

\begin{figure}[htbp]
  \centering
  \begin{tabular}{cccc}
    \begin{subfigure}{0.22\textwidth}
      \includegraphics[width=\linewidth]{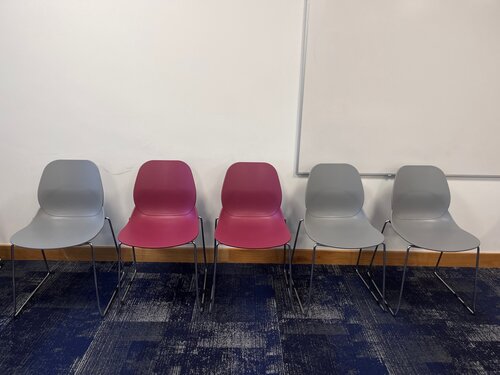}
      \caption{Inside}
    \end{subfigure} &
    \begin{subfigure}{0.22\textwidth}
      \includegraphics[width=\linewidth]{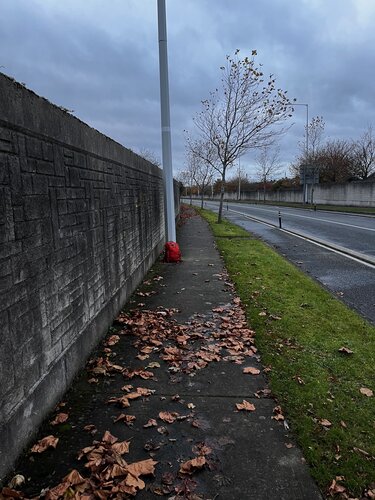}
      \caption{Outside}
    \end{subfigure} &
    \begin{subfigure}{0.22\textwidth}
      \includegraphics[width=\linewidth]{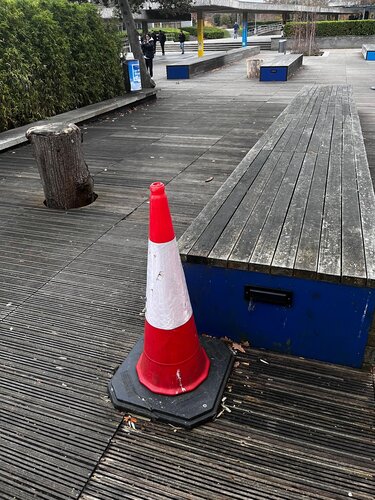}
      \caption{Day}
    \end{subfigure} &
    \begin{subfigure}{0.22\textwidth}
      \includegraphics[width=\linewidth]{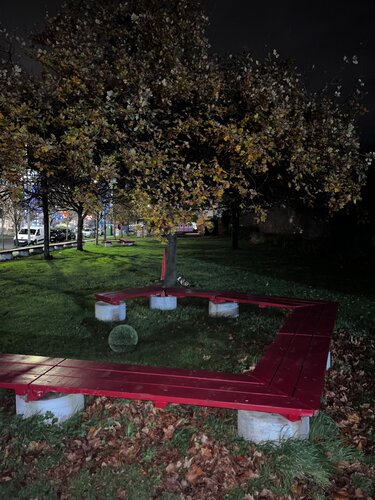}
      \caption{Night}
    \end{subfigure} \\
    
    \begin{subfigure}{0.22\textwidth}
      \includegraphics[width=\linewidth]{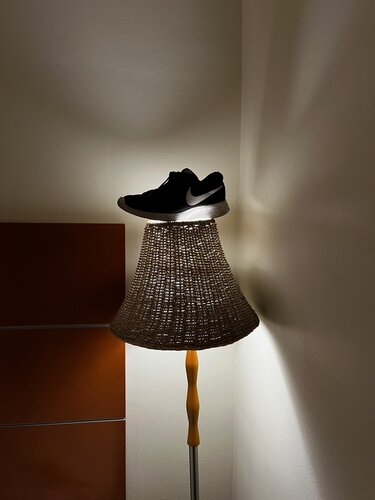}
      \caption{Artificial light}
    \end{subfigure} &
    \begin{subfigure}{0.22\textwidth}
      \includegraphics[width=\linewidth]{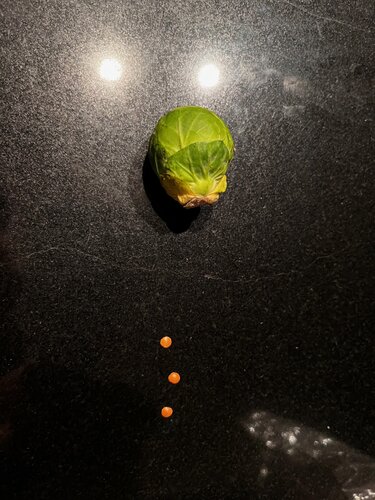}
      \caption{Reflections}
    \end{subfigure} &
        \begin{subfigure}{0.22\textwidth}
      \includegraphics[width=\linewidth]{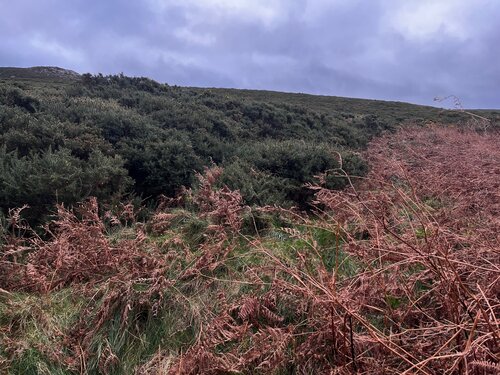}
      \caption{Natural environment}
    \end{subfigure} &
    \begin{subfigure}{0.22\textwidth}
      \includegraphics[width=\linewidth]{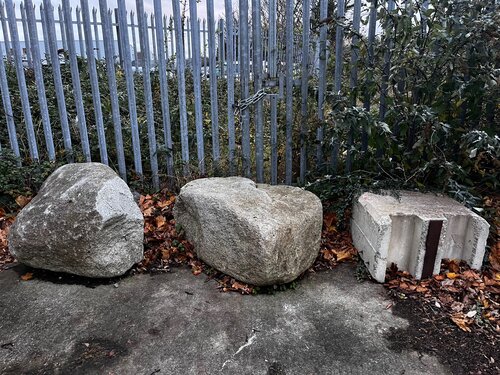}
      \caption{Urban environmen}
    \end{subfigure} \\


  \end{tabular}
  \caption{Examples for scene diversity in RocketScience.}
  \label{table:examples}
\end{figure}

\clearpage
\section{Appendix: Dataset Analysis}
\label{appendix:datasetanalysis}

\begin{figure}[h!] 
    \centering
    \includegraphics[width=0.8\textwidth]{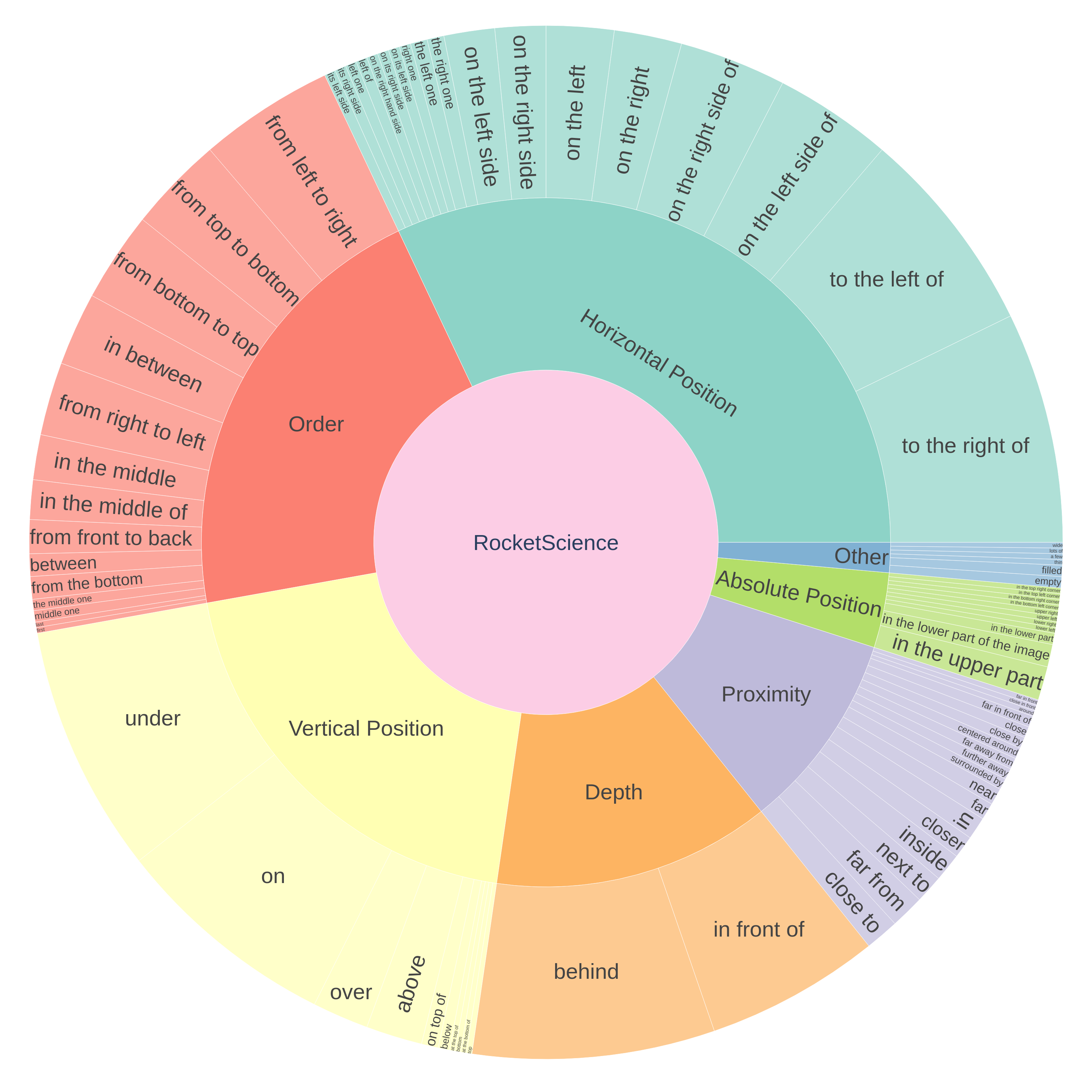}  
    \caption{Dataset distribution, relative proportions}
    \label{fig:datasetdistribution}
\end{figure}

\begin{figure}[h!]  
    \centering
    \includegraphics[width=0.75\textwidth]{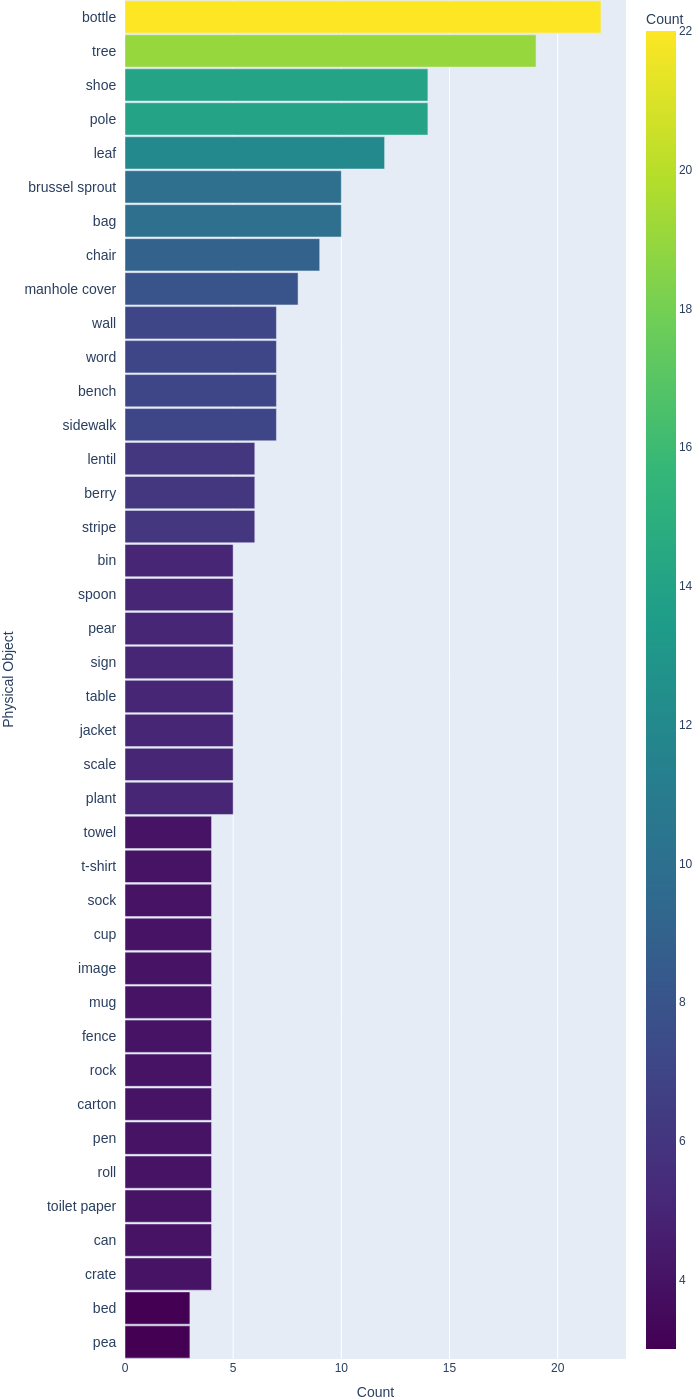} 
    \caption{Physical Object Counts}
    \label{fig:physicalobjectcounts}
\end{figure}

\begin{figure}[h!]  
    \centering
    \includegraphics[width=0.75\textwidth]{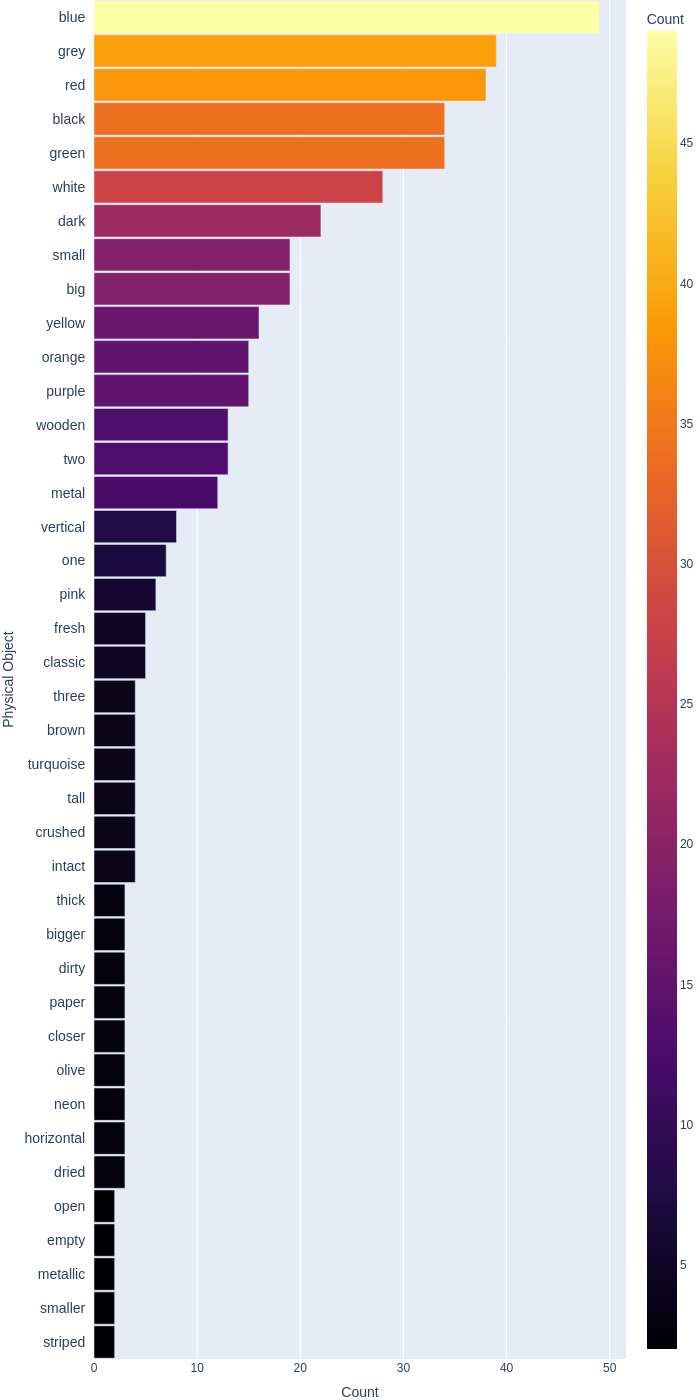}
    \caption{Adjective Counts}
    \label{fig:adjectivetcounts}
\end{figure}

\clearpage
\section{Appendix: Detailed Results}
\label{appendix:detailedresults}

\begin{table}[h]
    \centering
\begin{tabular}{lrrr}
\toprule
modelname & textscore & imagescore & groupscore \\
\midrule
random & 0.25 & 0.25 & 0.17 \\
human & 0.97 & 0.98 & 0.95 \\
\midrule
ViT-B-32negCLIP \cite{vl-bow}& 0.08 & 0.04 & 0.01 \\
EVA02-B-16merged2b\_s8b\_b131k \cite{evaclip} & 0.13 & 0.04 & 0.02 \\
EVA02-L-14-336merged2b\_s6b\_b61k \cite{evaclip} & 0.12 & 0.05 & 0.02 \\
ViT-B-16-SigLIPwebli \cite{siglip} & 0.13 & 0.04 & 0.02 \\
ViT-L-16-SigLIP-384webli \cite{siglip} & 0.10 & 0.07 & 0.02 \\
ViT-L-14-CLIPAdatacomp1b \cite{clipa}& 0.07 & 0.05 & 0.00 \\
ViT-L-16-SigLIP2-512webli \cite{siglip2} & 0.12 & 0.07 & 0.01 \\
coca\_ViT-B-32laion2b\_s13b\_b90k \cite{coca}& 0.14 & 0.05 & 0.02 \\
coca\_ViT-L-14laion2b\_s13b\_b90k \cite{coca} & 0.10 & 0.04 & 0.01 \\
ViT-B-16-SigLIP2-512webli \cite{siglip2}& 0.10 & 0.05 & 0.02 \\
ViT-B-16openai \cite{clip}& 0.11 & 0.05 & 0.02 \\
ViT-B-32openai \cite{clip}& 0.11 & 0.02 & 0.01 \\
\midrule
paligemma-3b-mix-448 \cite{paligemma} & 0.13 & 0.10 & 0.02 \\
qwen-2.5-vl-72b-instruct \cite{qwen2.5vl} & 0.47 & 0.01 & 0.00 \\
qwen-vl-max \cite{qwenvlmax} & 0.29 & 0.03 & 0.01 \\
claude-3-7-sonnet-20250219 \cite{claude3.7}& 0.53 & 0.37 & 0.24 \\
llama-4-maverick \cite{llama4} & 0.38 & 0.37 & 0.20 \\
gpt-4o-2024-08-06 \cite{gpt4o}& 0.38 & 0.39 & 0.19 \\
\midrule
llama-4-maverick\_cot \cite{llama4}& 0.59 & 0.66 & 0.44 \\
gpt-4o-2024-08-06\_cot \cite{gpt4o}& 0.73 & 0.66 & 0.51 \\
SpaceOm \cite{spaceOm} & 0.08 &  0.14 & 0.01\\
\midrule
glm-4.1v-9b-thinking \cite{glm41thinking} & 0.84 & 0.72 & 0.64 \\
gemini-2.5-pro-preview-03-25 \cite{gemini2.5} & \textbf{0.94} & 0.89 & 0.83 \\
o4-mini (medium) \cite{o4-mini} & 0.91 & \textbf{0.94} & \textbf{0.89} \\

\bottomrule
\end{tabular}
    \caption{Results on the RocketScience benchmark, the second division is CLIP-like models, the third regular VLMs, the fourth regular vlms with explicit chain-of-thought and the last VLMs with implicit chain-of-thought. All CLIP-like models and basic VLMs fail drastically while some reasoning models come very close to human performance.}
    \label{table:bigresults}
\end{table}

\begin{table}[h]
\centering
\begin{tabular}{l ccc ccc ccc}
\toprule
\textbf{Model} 
& \multicolumn{3}{c}{\textbf{Horizontal}} 
& \multicolumn{3}{c}{\textbf{Vertical}} 
& \multicolumn{3}{c}{\textbf{Depth}} \\
\cmidrule(lr){2-4} \cmidrule(lr){5-7} \cmidrule(lr){8-10}
& ts & is & gs          & ts & is & gs       & ts & is & gs                         \\
\midrule
paligemma-3b-mix-448 & 0.08 & 0.13 & 0.01 & 0.21 & 0.11 & 0.04 & 0.16 & 0.13 & 0.00 \\
qwen-2.5-vl-72b-instruct & 0.51 & 0.00 & 0.00 & 0.62 & 0.02 & 0.00 & 0.39 & 0.03 & 0.00 \\
qwen-vl-max & 0.23 & 0.07 & 0.01 & 0.47 & 0.02 & 0.02 & 0.24 & 0.00 & 0.00 \\
claude-3-7-sonnet-20250219 & 0.43 & 0.35 & 0.17 & 0.64 & 0.49 & 0.38 & 0.47 & 0.24 & 0.11 \\
llama-4-maverick & 0.40 & 0.33 & 0.17 & 0.45 & 0.36 & 0.23 & 0.18 & 0.29 & 0.05 \\
gpt-4o & 0.35 & 0.27 & 0.08 & 0.62 & 0.58 & 0.40 & 0.39 & 0.42 & 0.24 \\
llama-4-maverick\_cot & 0.67 & 0.72 & 0.53 & 0.55 & 0.66 & 0.42 & 0.37 & 0.50 & 0.21 \\
gpt-4o\_cot & 0.72 & 0.63 & 0.44 & 0.83 & 0.77 & 0.68 & 0.74 & 0.55 & 0.39 \\
gemini-2.5-pro-preview-03-25 & 0.99 & 0.92 & 0.91 & 0.96 & 0.91 & 0.87 & 0.89 & 0.84 & 0.76 \\
o4-mini & 0.97 & 0.97 & 0.97 & 0.92 & 0.94 & 0.91 & 0.95 & 0.92 & 0.89 \\
\bottomrule
\end{tabular}
\caption{Text score, image score and group score for each category in the dataset.}
\end{table}

\begin{table}[h]
\centering
\begin{tabular}{l ccc ccc ccc}
\toprule
\textbf{Model} 
& \multicolumn{3}{c}{\textbf{Proximity}} 
& \multicolumn{3}{c}{\textbf{Order}} 
& \multicolumn{3}{c}{\textbf{Absolute Position}} \\
\cmidrule(lr){2-4} \cmidrule(lr){5-7} \cmidrule(lr){8-10}
& ts & is & gs          & ts & is & gs       & ts & is & gs                         \\
\midrule
paligemma-3b-mix-448 & 0.04 & 0.00 & 0.00 & 0.13 & 0.07 & 0.02 & 0.40 & 0.60 & 0.20 \\
qwen-2.5-vl-72b-instruct & 0.44 & 0.04 & 0.00 & 0.37 & 0.00 & 0.00 & 0.60 & 0.00 & 0.00 \\
qwen-vl-max & 0.44 & 0.00 & 0.00 & 0.23 & 0.03 & 0.02 & 0.20 & 0.00 & 0.00 \\
claude-3-7-sonnet-20250219 & 0.72 & 0.68 & 0.52 & 0.53 & 0.28 & 0.18 & 0.40 & 0.00 & 0.00 \\
llama-4-maverick & 0.64 & 0.64 & 0.52 & 0.25 & 0.37 & 0.15 & 0.80 & 0.40 & 0.40 \\
gpt-4o & 0.52 & 0.60 & 0.40 & 0.18 & 0.22 & 0.07 & 0.20 & 0.40 & 0.00 \\
llama-4-maverick\_cot & 0.68 & 0.88 & 0.64 & 0.68 & 0.57 & 0.45 & 0.40 & 0.80 & 0.20 \\
gpt-4o\_cot & 0.68 & 0.76 & 0.56 & 0.72 & 0.60 & 0.52 & 0.60 & 1.00 & 0.60 \\
gemini-2.5-pro-preview-03-25 & 0.88 & 0.88 & 0.76 & 0.92 & 0.87 & 0.80 & 1.00 & 0.80 & 0.80 \\
o4-mini & 0.72 & 0.88 & 0.68 & 0.90 & 0.92 & 0.88 & 1.00 & 1.00 & 1.00 \\
\bottomrule
\end{tabular}
\caption{Textscore, imagescore and groupscore for each category in the dataset}
\end{table}

\clearpage
\section{Appendix: Evaluation Stability}
\label{appendix:stability}
To prove that the size of our benchmark is appropriate, we test the standard deviation of one model with
poor performance (gpt4o without chain-of-thought) and one model with good performance (gemini 2.5 pro). We run each model three times and then randomly sample subsets of size 0.5 to 1.0 of
the dataset and provide their mean and standard deviation below. RocketScience yields stable evaluation results and would even do so if it were much smaller.

\begin{table}[h!]
\centering
\begin{tabular}{ccc}
\hline
\textbf{Share} & \textbf{Gpt-4o (Mean $\pm$ Std)} & \textbf{Gemini 2.5 pro (Mean $\pm$ Std)} \\
\hline
0.5 & 0.21 $\pm$ 0.03 & 0.86 $\pm$ 0.01 \\
0.6 & 0.20 $\pm$ 0.03 & 0.85 $\pm$ 0.01 \\
0.7 & 0.20 $\pm$ 0.03 & 0.85 $\pm$ 0.01 \\
0.8 & 0.19 $\pm$ 0.02 & 0.84 $\pm$ 0.01 \\
0.9 & 0.19 $\pm$ 0.02 & 0.85 $\pm$ 0.02 \\
1.0 & 0.18 $\pm$ 0.02 & 0.86 $\pm$ 0.02 \\
\hline
\end{tabular}
\label{stability}
\caption{Performance of Gpt-4o and Gemini 2.5 pro over three runs each on random subsets of RocketScience. The standard deviation stays low at both half the dataset size and the full dataset.}
\end{table}

\clearpage
\section{Appendix: Human Baseline}
\label{appendix:humanbaseline}

The full set of instructions given to the participants (apart from the consent form) is: "You will be asked to answer several questions. Each question will consist of two images and a caption, and you will need to click on the image that best matches the caption." 

\begin{figure}[h!]  
    \centering
    \includegraphics[width=0.75\textwidth]{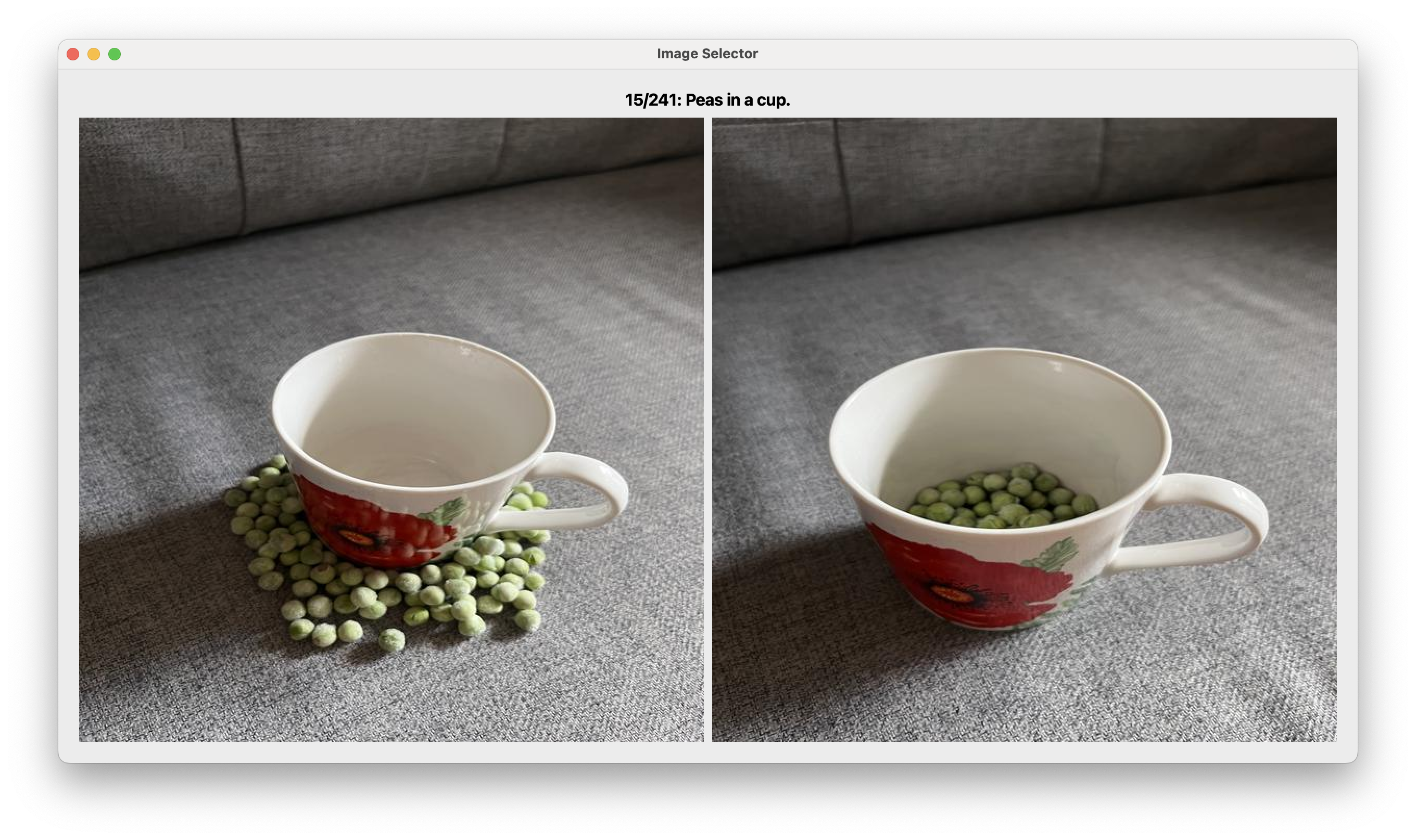}
    \caption{Human baseline interface with two images}
    \label{figure:twoimages}
\end{figure}

\begin{figure}[h!]  
    \centering
    \includegraphics[width=0.6\textwidth]{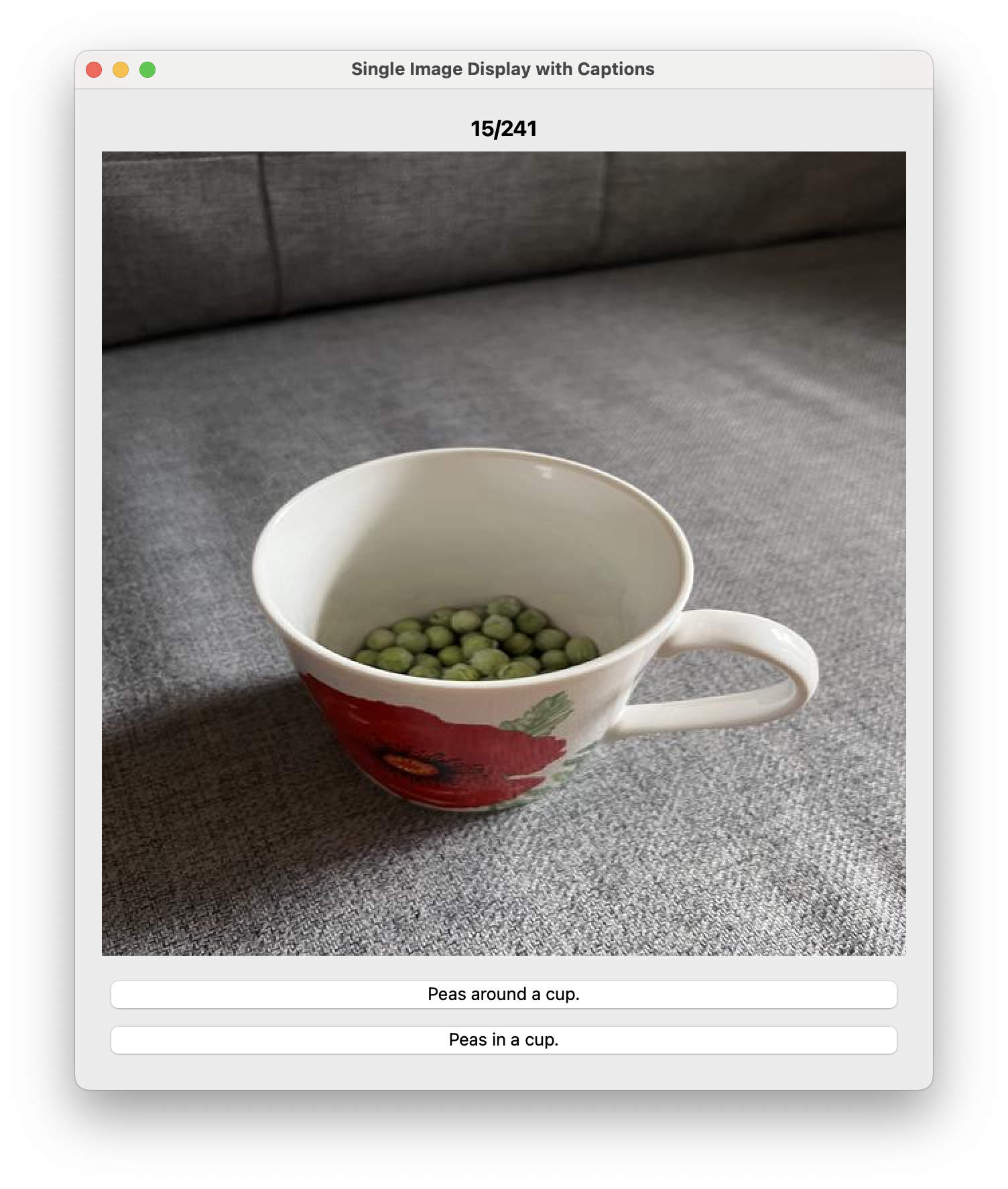}
    \caption{Human baseline interface with two captions}
    \label{figure:twocaptions}
\end{figure}

The testing interface can be seen in Figure \ref{figure:twoimages} and Figure \ref{figure:twocaptions}.

We do not see any significant risks for the study participants and we obtained permission for the human evaluation from University College Dublin's Human Research Ethics Committee.

\newpage
\section*{NeurIPS Paper Checklist}

\begin{enumerate}

\item {\bf Claims}
    \item[] Question: Do the main claims made in the abstract and introduction accurately reflect the paper's contributions and scope?
    \item[] Answer: \answerYes{}
    \item[] Justification: 1) we provide an overview of existing spatial relations benchmarks and their shortcomings in the related work section, 2) we propose RocketScience, a manually curated benchmark that addresses the shortcomings in the main section 3) We examine whether the localization or reasoning is more important for the success of CoT models in the results section.

\item {\bf Limitations}
    \item[] Question: Does the paper discuss the limitations of the work performed by the authors?
    \item[] Answer: \answerYes{} 
    \item[] Justification: We include a section on limitations and also discuss limitations throughout the rest of the text when necessary.

\item {\bf Theory assumptions and proofs}
    \item[] Question: For each theoretical result, does the paper provide the full set of assumptions and a complete (and correct) proof?
    \item[] Answer: \answerNA{} 
    \item[] Justification: We do not include theoretical results.

    \item {\bf Experimental result reproducibility}
    \item[] Question: Does the paper fully disclose all the information needed to reproduce the main experimental results of the paper to the extent that it affects the main claims and/or conclusions of the paper (regardless of whether the code and data are provided or not)?
    \item[] Answer: \answerYes{} 
    \item[] Justification: The dataset is published on huggingface and the evaluation script is available on github via the link in the abstract, including instructions for how to run it. The model's full names are available in the code, but also stated in the results tables for full reproducibility.

\item {\bf Open access to data and code}
    \item[] Question: Does the paper provide open access to the data and code, with sufficient instructions to faithfully reproduce the main experimental results, as described in supplemental material?
    \item[] Answer: \answerYes{} 
    \item[] Justification: The dataset is published on huggingface and the evaluation script is available on github via the link in the abstract, including instructions for how to run it. The model's full names are available in the code, but also stated in the results tables for full reproducibility.

\item {\bf Experimental setting/details}
    \item[] Question: Does the paper specify all the training and test details (e.g., data splits, hyperparameters, how they were chosen, type of optimizer, etc.) necessary to understand the results?
    \item[] Answer: \answerYes{} 
    \item[] Justification: We describe all hyperparameters used for evaluation of the models and they are also available in the code.

\item {\bf Experiment statistical significance}
    \item[] Question: Does the paper report error bars suitably and correctly defined or other appropriate information about the statistical significance of the experiments?
    \item[] Answer: \answerYes{} 
    \item[] Justification: We report mean and standard deviation in our experiment in table 3a). For our main results table we explain how we achieve reproducibilty on open models and that we can only run API models once due to their cost.

\item {\bf Experiments compute resources}
    \item[] Question: For each experiment, does the paper provide sufficient information on the computer resources (type of compute workers, memory, time of execution) needed to reproduce the experiments?
    \item[] Answer: \answerYes{} 
    \item[] Justification: We point out the type of GPU and rough execution times.
    
\item {\bf Code of ethics}
    \item[] Question: Does the research conducted in the paper conform, in every respect, with the NeurIPS Code of Ethics \url{https://neurips.cc/public/EthicsGuidelines}?
    \item[] Answer: \answerYes{} 
    \item[] Justification: We have a section on ethics where we discuss how we avoid potential issues.

\item {\bf Broader impacts}
    \item[] Question: Does the paper discuss both potential positive societal impacts and negative societal impacts of the work performed?
    \item[] Answer: \answerYes{} 
    \item[] Justification: We discuss societal impacts in the ethics section.
    
\item {\bf Safeguards}
    \item[] Question: Does the paper describe safeguards that have been put in place for responsible release of data or models that have a high risk for misuse (e.g., pretrained language models, image generators, or scraped datasets)?
    \item[] Answer: \answerNA{} 
    \item[] Justification: Our dataset does not contain personal information and we are not aware of high-risk scenarios posed by the release.

\item {\bf Licenses for existing assets}
    \item[] Question: Are the creators or original owners of assets (e.g., code, data, models), used in the paper, properly credited and are the license and terms of use explicitly mentioned and properly respected?
    \item[] Answer: \answerYes{} 
    \item[] Justification: We cite all models used and for images that are not ours we cite the source and state the license of the image.

\item {\bf New assets}
    \item[] Question: Are new assets introduced in the paper well documented and is the documentation provided alongside the assets?
    \item[] Answer: \answerYes{} 
    \item[] Justification: We document the details of our dataset in the paper as well as on huggingface and also submit a croissant file.

\item {\bf Crowdsourcing and research with human subjects}
    \item[] Question: For crowdsourcing experiments and research with human subjects, does the paper include the full text of instructions given to participants and screenshots, if applicable, as well as details about compensation (if any)? 
    \item[] Answer: \answerYes{} 
    \item[] Justification: We provide full documentation of the human performance experiment in the appendix.

\item {\bf Institutional review board (IRB) approvals or equivalent for research with human subjects}
    \item[] Question: Does the paper describe potential risks incurred by study participants, whether such risks were disclosed to the subjects, and whether Institutional Review Board (IRB) approvals (or an equivalent approval/review based on the requirements of your country or institution) were obtained?
    \item[] Answer: \answerYes{} 
    \item[] Justification: We obtained permission for this experiment by University College Dublin's Human Research Ethics Committee as a low-risk study.

\item {\bf Declaration of LLM usage}
    \item[] Question: Does the paper describe the usage of LLMs if it is an important, original, or non-standard component of the core methods in this research? Note that if the LLM is used only for writing, editing, or formatting purposes and does not impact the core methodology, scientific rigorousness, or originality of the research, declaration is not required.
    \item[] Answer: \answerNA{} 
    \item[] Justification: The core contribution is a benchmark which was manually created.

\end{enumerate}

\end{document}